%% file: main.tex
\documentclass[review]{elsarticle}
\usepackage{comment}
\includecomment{comment} 
\usepackage{lineno,hyperref}
\usepackage{hyperref}
\usepackage{multirow}
\usepackage{longtable}
\usepackage{lscape}
\usepackage{graphicx}
\usepackage{subcaption}
\usepackage{amssymb, amsmath, bm}
\usepackage[utf8]{inputenc}
\usepackage{mathtools}
\usepackage{url}
\usepackage{color, soul}
\sethlcolor{green}

\usepackage{xcolor}

\def\Tiny{\fontsize{4pt}{4pt}\selectfont}
\newcommand*{\eqdef}{\ensuremath{\overset{\mathclap{\text{\Tiny def}}}{=}}}

\DeclareMathOperator*{\argminA}{argmin}

\nolinenumbers 

\journal{Biochemical Engineering Journal}

\begin{document}

\begin{frontmatter}

%\title{Elsevier \LaTeX\ template\tnoteref{mytitlenote}}
\title{When Bioprocess Engineering Meets Machine Learning: A Survey from the Perspective of Automated Bioprocess Development}

% Reproducability no longer in the title: Still mention it in relation to FAIR principles.


%\author{Nghia Duong-Trung, Stefan Born}

\author[mymainaddress]{Nghia Duong-Trung\corref{mycorrespondingauthor}}
\author[mymainaddress]{Stefan Born}
\author[mymainaddress]{Jong Woo Kim}
\author[mymainaddress]{Marie-Therese Schermeyer}
\author[mymainaddress]{Katharina Paulick}
\author[mymainaddress]{Maxim Borisyak}
\author[mymainaddress]{Mariano Nicolas Cruz-Bournazou}
\author[mysecondaryaddress]{Thorben Werner}
\author[mysecondaryaddress]{Randolf Scholz}
\author[mysecondaryaddress]{Lars Schmidt-Thieme}
\author[mymainaddress]{Peter Neubauer}
\author[mymainaddress,mythirdaddress]{Ernesto Martinez\corref{mycorrespondingauthor}}

\cortext[mycorrespondingauthor]{Corresponding authors: Nghia Duong-Trung and Ernesto Martinez}




\address[mymainaddress]{Technische Universität Berlin, Faculty III Process Sciences, Institute of Biotechnology, Chair of Bioprocess Engineering.\\Straße des 17. Juni 135, 10623 Berlin, Germany.}
\address[mysecondaryaddress]{University of Hildesheim, Information Systems and Machine Learning Lab (ISMLL).\\Universitätspl. 1, 31141 Hildesheim, Germany.}
\address[mythirdaddress]{INGAR (CONICET-UTN), Avellaneda 3657, S3003GJC, Santa Fe, Argentina.}
%\address[mythirdaddress]{360 Park Avenue South, New York}

\begin{abstract}
%Machine learning (ML) has significantly contributed to the development of bioprocess engineering, but its application is still limited, hampering the enormous potential for bioprocess automation.
%
% I hope the following does not sound too negative,
% but I think that is the story, that up to now ML
% is not a predominant tool in bioprocess development. SB
%
Machine learning (ML) is becoming increasingly crucial in
many fields of engineering but has not yet played
out its full potential in bioprocess engineering.
While experimentation has been
accelerated by increasing levels of lab automation,
experimental planning and data modeling are still largerly
depend on human intervention.
ML can be seen as a set of tools that contribute to 
the automation of the whole experimental cycle, including 
model building and practical planning,
thus allowing human experts to focus on the more
demanding and overarching cognitive tasks. 
%ML for model building automation is a way of introducing another level of abstraction to focus expert humans on the most cognitive tasks of bioprocess development. 
%
%  I'll have a look at the abstract again after working
% on the rest of the article.
%
First, probabilistic programming is used for the autonomous building of predictive models. 
Second, machine learning automatically assesses alternative decisions by planning experiments to test hypotheses and conducting investigations to gather informative data that focus on model selection based on the uncertainty of model predictions.
This review provides a comprehensive overview of ML-based automation in bioprocess development. 
On the one hand, the biotech and bioengineering community should be aware of the potential and, most importantly, the limitation of existing ML solutions for their application in biotechnology and biopharma. 
On the other hand, it is essential to identify the missing links to enable the easy implementation of ML and Artificial Intelligence (AI) tools in valuable solutions for the bio-community.
%We summarize recent ML deployment across several important subfields of bioprocess engineering and raise two crucial challenges that are still pending: the bottleneck of insufficient automation and the need for systematically reducing uncertainty in bioprocess development by gathering informative data using active learning.
There is no one-fits-all procedure; however, this review should help identify the potential automation combining biotechnology and ML domains.
\end{abstract}

\begin{keyword}
Active Learning \sep Automation \sep Bioprocess Development \sep Reinforcement Learning \sep Transfer Learning.
%\texttt{elsarticle.cls}\sep \LaTeX\sep Elsevier \sep template
%\MSC[2010] 00-01\sep  99-00
\end{keyword}

\end{frontmatter}

%\linenumbers

\section{Introduction}
\label{Introduction}
\input{tex/Introduction-new.tex}

%\section{Towards Automated Model Building in  Bioprocess Development via Machine Learning}
%\label{Motivations}
%\input{tex/Motivations-new.tex}

\section{Elucidation of Machine Learning Strategies}
\label{Elucidation}
\input{tex/Elucidation.tex}

\section{Current Integration of Machine Learning in Bioprocess Subfields}
\label{State-of-the-art}
\input{tex/State-of-the-art.tex}

\section{Challenges and Future Research Directions}
\label{Challenges}
\input{tex/Challenges.tex}

\section{Conclusion}
\label{conclusion}
\input{tex/Conclusion.tex}

%\section{Broader Impact}
%\label{impact}
%\input{tex/Impact.tex}

\section{Acknowledgments}
The authors kindly appreciate the support of the German Federal Ministry of Education and
Research through the Program "International Future Labs for Artificial
Intelligence (Grant number 01DD20002A)".
We acknowledge the Open Access Publication Fund of Technische Universität Berlin.

%\section*{References}

\bibliography{mybibfile}

\end{document}

%% file: tex/Introduction-new.tex
In the wake of climate change, many industries are turning to biotechnology to find sustainable solutions. The importance of biotechnological processes in pharmaceuticals is reflected in the growth figures for biopharmaceuticals (up 14 \% to 30.8 \% market share from 2020 to 2021)\cite{lucke_medizinische_2021}. 
This trend is currently strongly inhibited by long development times of biotechnological processes. To advance fast in bioprocess development, decisions must be taken under considerable high uncertainty which does not enable a fast transition from laboratory to industrial production at scale with acceptable risks. Usually, different microorganisms or cells are tested to produce an industrial relevant product, i.e., a pharmaceutical substance. The transfer of results from small to large scale represents a central challenge and is very time-consuming and error-prone. 

Modern biolabs have automatized and parallelized many tasks aiming to run such a large number of experiments in short periods of time. These Robotic experimental facilities are equipped with Liquid Handling Stations (LHS) \cite{waldbaur2013microfluidics,radtke2016photoinitiated}, parallel cultivation systems, and High Throughput (HT) \cite{treier2012high,reuter2015high} analytical devices which make them capable of timely generating informative data over a wide range of operating conditions. 
The past decade's focus was on hardware development and device integration with fairly simple data management systems lacking automatic association of the relevant metadata for the resulting experimental data. We have not been able to trigger the fruitful symbiosis expected between i. robots that can perform thousands of complex tasks but are currently waiting for humans to design and operate the experiments, ii. Active Learning (AL) algorithms that still rely on humans to perform the planned experiments, and iii. Machine Learning (ML) tools that are at the present waiting for humans to treat and deliver the data in a digital, machine-actionable format. Hence, end-to-end digitalization of experiments is a pre-requisite to apply ML methods in bioprocessing.

Without complete annotations, the knowledge about how data were actually generated remains hidden,  thus limiting the possible  degree of automation for control and experimental design, but also hampering the aggregation of data from different contexts. More importantly, difficulties to reproduce experiments prevents sharing and reuse by other researchers of costly experimental data. 

Accordingly, with the advent of high-throughput robotic platforms, the bottleneck to efficient experimentation on a micro scale has thus shifted from running a large number of parallel experiments to data management, model building and experimental design,  all of which currently rely on a considerable amount of human intervention which makes experiments barely reproducible. Only a proper data management system with standardized machine-actionable data and automated metadata capture would allow an automatic flow of information through all stages of experimentation in bioprocess development and facilitate the use of machine learning models for decision making in the face of uncertainty.

As a representative example of the importance of metadata and experiment reproducibility, let us consider scale-up. Miniaturized and versatile multi-bioreactor systems combined with LHS have the potential to significantly contribute to the effective generation of informative data to increase scale-up efficiency bearing in mind robustness to face the variability in operating conditions during strain selection at the initial stage. When transferring the acquired knowledge in the lab to the  industrial scale the remaining uncertainty in model predictions are significantly high due to insufficient data annotation and low levels of automation. Hence, key decisions must be taken under high uncertainty which imposes significant risks to most decisions taken throughout the bioprocess lifecycle. 

The different stages of the development cannot be treated in isolation. For example,
the variability of  operating conditions during strain selection has a direct influence on the reproducibility of productivity levels in the scaled process. Hence, a promising route to faster development of innovative bioprocesses is a comprehensive automation of model building and experimental design across all stages
of the development. To drastically speed up the bioprocess development of innovative products, the ubiquitous use of automation in active learning from data and model building must be introduced in all stages from product conceptualization to reproducible end-use properties. At any of these development stages, problem-solving and decision-making require building a model with enough predictive capability as well as a proper evaluation of its associated uncertainty. 

In today's practice, model building and data collection depend heavily on manual tweaking and human intervention, which slows down the development effort and constitute a significant obstacle to lower costs and shorter times to market. Also, ML algorithms should be deployed with higher levels of autonomy to release the end-user from choosing among alternatives for algorithms, hyper-parameters and problem representation which are not compatible with their understanding of the methods involved and underlying assumptions.

This review follows two crucial aims. On the one hand, the biotech and bioengineering community should be aware of the potential and, most importantly, some limitations of existing ML methods for their application in biotechnology and biopharma. On the other hand, it is essential to identify the missing links to enable the easy implementation of ML and Artificial Intelligence (AI) solutions in valuable solutions for the bio-community as end-users.To drastically speed up bioprocess development of innovative products, the ubiquitous use of automation in active learning from data and model building must be introduced in all stages from product conceptualization to reproducible end-use properties. At any of these development stages, problem-solving and decision making requires building a model with enough predictive capability to assess the costs at stakes and risks involved.  

\subsection{Decisions and Models}

\begin{figure}
\includegraphics[width=0.8\textwidth]{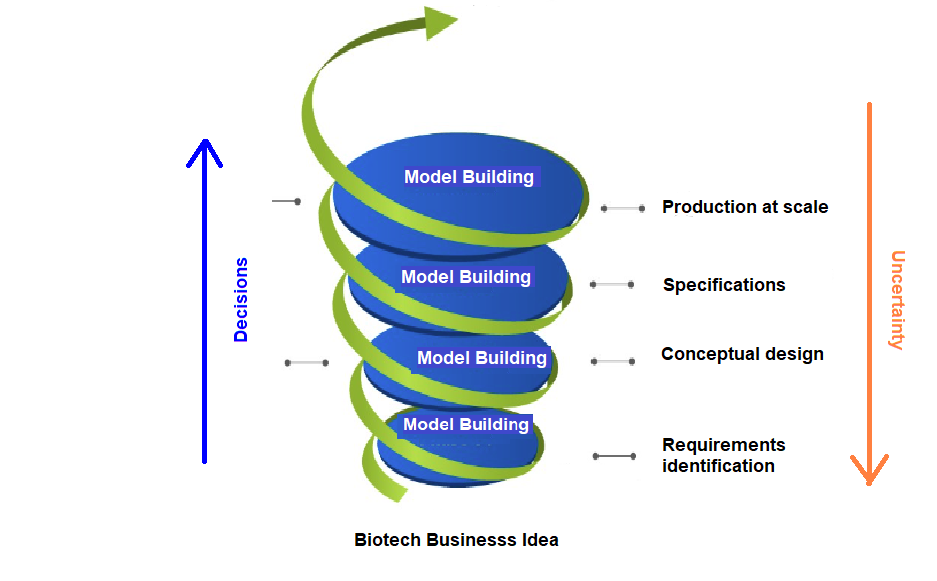}
\caption{Reducing uncertainty in the bioprocess development lifecycle.}
\label{spiral}
\end{figure}

As shown in Fig. \ref{spiral}, model building is an important activity to assess alternatives and advance fast in the bioprocess lifecycle by making rational decisions that systematically reduce uncertainty. Model-based decision-making is widely used in the development lifecycle of different types of processes and products (e.g., electrical, chemical, aeronautics) for cost-effective design and improved operation in the face of uncertainty. Mainly due to the so-called "small data problem," \cite{rogers2022} bioprocess development has been an exception, though, with a significantly higher degree of empiric procedures, expert-based decisions, and strongly segmented design strategies and strain screening methods. The increased complexity of living organisms with thousands of intracellular biochemical reactions and uncomprehended responses in its metabolic activity due to regulatory mechanisms, combined with the difficulties in obtaining trustworthy observations, makes it very difficult to build sound mathematical models since data collected from biological systems are inherently scarce and low‐dimensional. However,  similar dynamic behaviors within families of genetically modified microorganisms make enough room for transfer learning and meta-learning, using available data (with their metadata) to build predictive models for a new, unseen mutant. Based on such prior knowledge, experiments can be readily designed to gather informative data to increase the predictive power of models built to effectively support decision making.

\subsection{Automated Model-Building}

Automation of the model-building cycle aims to assist experts and scientists in facilitating and transforming decision-making in the context of bioprocess engineering and biotechnology, not replacing them. Some aspects of model-building are more difficult to automate because of technological challenges and because they involve open-ended questions and context-dependent tasks that require human cognitive abilities. Perhaps the most difficult challenge to model-building automation is that data sources in the development pipeline are diverse, distributed, and multi-structured. Moreover, not only the available data is highly heterogeneous, but also they may be not informative enough regarding the purpose of the model for a given stage. This fact contrasts sharply with the common assumption that sufficient amounts of high-quality data are available for model building. Data collection methods for bioprocess development are descriptive of an inherently dynamic behavior since many process parameters are gathered online as continuous measurements. They are available and highly dependent on the quality of sensor devices, standard operating procedures, material and methods used, frequency of measurements (e.g., temperature, optical density, pH, oxygen, glucose, oxygen uptake rate, stirring speed), analytical sample processing, device calibration parameters, etc. Then, the optimal design of experiments for gathering information must be an integral part of the model-building cycle. As a result, the model-building at the different stages of bioprocess development will comprise automated procedures for actively seeking or generating highly informative data regarding the objective and type of decisions that must be taken in each stage in Fig. \ref{spiral}. This is related to the machine learning
subfield known as 'active learning' (AL). As an example, data which are informative for strain screening highlights the robustness of different strains to scale up effects where lack of enough aeration influences the physiological state of the bioreactor. However, these data are possibly poorly informative on how to optimize the productivity of the chosen strain after scale up.

Faced with the choice of a large set of machine learning algorithms and an even larger space of hyperparameter settings, experts often must resort to costly experimentation in time and money to determine what combination works best for a given problem. Hence, automated model-building approaches must include automatic model selection, hyperparameter tuning, model training  and model validation. If possible, models should be trained and validated with respect to the final properties of interest based on an end-to-end approach. This not only spares non-experts the time and effort of extensive, often onerous trial-and-error experimentation but also enables bioprocess engineers to obtain substantially better performance with fewer data and faster than possible without automation. In some applications of reinforcement learning to control and optimize, close-loop experimentation must be part and parcel of the model building cycle, making automation even more important. Hyperparameters, in this case, drive learning and define which data is gathered in the learning curve. Without a meta-learning level in model building, the initial setting of hyperparameters can easily prevent learning a predictive model that can make a real difference compared to not using a model at all for taking decisions.
The use of ML for model-building automation can be seen as a way of introducing another level of abstraction that allows human experts to focus  on higher level cognitive tasks for bioprocess development. First, probabilistic programming is used for the autonomous building of predictive models. Second, ML automatically assesses alternative decisions by planning experiments to test hypotheses and then planning and executing experiments to gather informative data that focus on model selection based on the uncertainty of model predictions. Therefore, ML methods can be seen as meta-algorithms for model-building tasks and automated data generation and hypothesis testing. Finally, the automated model-building uses algorithms that select and configure ML algorithms. That is, meta-meta-algorithms that can be understood as Bayesian machine experimenters that can generate autonomously new data to transform a priori knowledge into rational decisions that further bioprocess development.

\subsection{Present State of Data and Models in Bioprocess Development}

At the initial stages of development (see \ref{spiral}), fundamental problems are addressed and key decisions are taken, such as strain screening which involves testing their robustness to alternative operating conditions, cultivation media, and bioreactor designs. The availability of Process Analytical Tools (PAT) \cite{kansakoski2006process,glassey2011process,simon2015assessment} allows a deeper understanding of the processes and the technological advances of HT and LHS in robotic platforms \cite{diederich2017high,barz2018adaptive,hans2018automated} that can generate large amounts of experimental data to feed the model-building cycle. Yet, the bottleneck step of human-in-the-loop prevents a rapid transition toward design and operating decisions at larger scales. An essential link is missing toward model-based bioprocess systems engineering \cite{koutinas2012bioprocess}: the conversion of automated experimental tasks and data (e.g., cultivation, sampling, analytics) into knowledge expressed in mathematical expressions. The large amounts of heterogeneous low quality data make manual treatment and model development almost impossible. Automating model-building using ML is envisioned as the alternative of choice to speed up automated bioprocess development while providing a setting for provenance and reproducibility to transform HT experimental data into information, information into knowledge, and to use this knowledge to understand, control, and optimize the bioprocess throughout its entire lifecycle.

Machine learning tools are already contributing to accelerated drug discovery \cite{dara_machine_2022} and have the potential also to speed up process development for biopharmaceuticals. 
When the ML tools are used in the actual production of pharmaceuticals, the requirements of regulatory bodies (e.g., FDA) for 
good manufacturing practices and process performance qualifications become an issue.  
In the context of software as a medical device (SaMD), the FDA published a paper on a proposed regulatory framework \cite{food2019proposed}. 
A database of FDA-approved SaMD applications until 2020 contained 
64 medical applications based on ML/AI \cite{benjamens_state_2020}, but notably, only 29 of the items used machine learning or artificial intelligence-related terms in the official FDA documents. 
The FDA used to validate 'locked' algorithms only, that is, algorithms with parameters after training such that the same input would always map to the same output. 
Fortunately, the proposed regulatory framework shows that the FDA knows that many or perhaps the most relevant machine learning applications would be adaptive and continuously retrained on new data. 
Instead of a fixed input-output behavior, this requires a total product lifecycle regulatory approach, which determines how exactly models are retrained and validated. 
How far this approach will determine the FDA's behavior towards ML in manufacturing remains to be seen.
The uncertainty that reigns until an explicit statement by the FDA and other regulatory bodies may, at present, deter companies from using ML in production. 
But, as we have seen, SaMD devices based on ML, which were not declared as such, have been validated by the FDA. 
The same may apply to process analytical components that are packaged as soft sensors but rely on ML.

As the developmental stages are more concerned with decisions related design and operating conditions, the model-building should focus on guaranteeing physiological conditions that maximize productivity and product quality. For example, in the fed-batch cultivation phase, both overfeeding and underfeeding typically yield inferior results in cell growth and product formation \cite{lee1999control}. Several studies have resorted to mechanistic models for (re)designing HT experiments of several fed-batch mini-bioreactors. The main challenges which are pending to be addressed are i) the use of impulsive control systems due to bolus-feeding for a miniaturized system, ii) ill-conditioned parameter estimation, and iii) low predictive power of mechanistic models.

In the work of \cite{cruz2017online, barz2018adaptive, kim2021oed}, optimal experimental design problems were studied to maximize the information content for effective identification of a mechanistic model. Model predictive control using a mechanical model to maximize cell growth was implemented to an in silico system \cite{kim2022model} and validated using an HT experiment \cite{krausch2022high}.

However, due to its imperfect structure, a mechanistic model alone cannot significantly reduce the uncertainty related to operating and design decisions at more advanced stages. The latest trend clearly shows that machine learning techniques may give more room for more efficient utilization of available data and automate the generation of highly informative new data. Machine learning can provide  viable and effective solution to the preceding problems.  
Over the past few decades, biotechnology has seen a significant shift from manual modeling to data-driven modeling, e.g., applying ML, partly due to a large amount of existing data for some biological systems \cite{mowbray2021machine,lawson2021machine,scheper2020digitalization}. 
It is an essential premise for integrating machine learning models that are built based well-informed bio-data which are both FAIR and comprehensively annotated. Thus, data-driven models offer an appealing alternative for autonomous discovering in the field of bioprocessing \cite{narayanan2020bioprocessing,neubauer2017bioprocess,neubauer2020potential}. 

Data-driven models are validated by their performance on the tasks for which they are trained.  We should, however, bear in mind that
 unlike models built on first principles models learned from data will only extrapolate well to data coming from the same distribution. This may be the reason
 for a certain reluctance to adopt machine learning models, seen as black boxes
 without an interpretation. (It is, of course,
 possible to analyze or explain a machine learning model, once it is trained.) 
 And yet, whenever no satisfactory  first principles model is available, machine learning is the method of choice despite its lack of interpretability.
 
Machine learning has proved its effectiveness in many areas of biology: 3D structure of proteins \cite{wei2019protein,jumper2021highly}, up-downstream processes \cite{kaspersetz2022automated,schonberger2018deep,haque2016artificial,walther2022smart}, bioprocessing for chemical and biologic product manufacturing \cite{scheper2020digitalization,petsagkourakis2020reinforcement}, enzymes and cell growth \cite{mazurenko2019machine,heckmann2018machine,tan2019survey}, cell culture expression systems \cite{barz2022characterization,borisov2017individual,ashraf2021applying}, and many others. However,According to \cite{venkatasubramanian2019promise}, machine learning has not been as extensively used for bioprocess development as one might expect.
This may be attributed to various reasons, of
which some have already been mentioned. Data
management and curation with an appropriate ontology for the metadata is one requirement not yet met. Furthermore the "small data problem" makes the off-the-shelf use of 
existing models problematic.  Models have to be specially tailored to  be expressive enough for the complexity of the investigated phenomena, but constrained enough to be
trainable with the available data. Overarching data management standards may also help to aggregate data and to tackle the small data problem from the opposite side.  
But even with appropriate models and well managed data,  model selection, hyperparameter tuning, training and validation are still cognitive demanding tasks. Automation of this model building cycle is mandatory to increase the adoption of machine learning tools in 
bioprocess engineering.
The selection of the most efficient algorithm and its parameters is based on many factors, including the transformation from a bioprocessing engineering problem into machine learning tasks, the quantity and quality of the data collected,  the type of problem being solved (regression, classification,forecasting, control, etc.), the required overall accuracy and performance, availability of prior bioprocessing knowledge to control the hyperparameters tuning \cite{niazi2017fundamentals,neubauer2017continuous}. As a result, a key challenge for model building automation is integrating a meta-learning layer for setting all hyper-parameters using techniques such as Bayesian optimization, which can take full advantage of cumulative data in the bioprocess lifecycle to systemically reduce uncertainty.

.

%% file: tex/Elucidation.tex
%\section{Elucidation}

\subsection{Key Concepts}
\label{sec:key-concepts}

\subsubsection{Brief Definition of Machine Learning in the Context of Bioprocess Engineering}
\label{subsec:machine-learning}

 Machine learning is a field of computer science and statistics that deals with data-driven modelling and algorithms. It thus can be seen as a new form of computational statistics applicable when no explicit mathematical description of relationships between data is known from theory.  Machine learning has been particularly successful in domains where large amounts of data with a complex structure can     be aggregated.

\begin{figure}[ht!]
	\centering
	\includegraphics[width=0.99\linewidth]{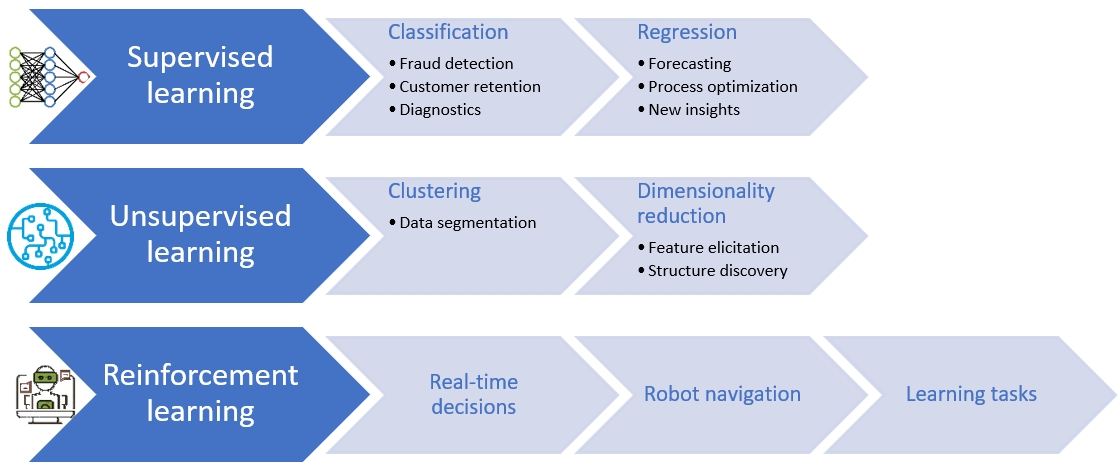} %
	\caption{Sub-fields of machine learning.}
	\label{fig:ml-subfields}
\end{figure}

% Machine learning is a subfield of computer science that focuses on building applicable models and depends on training upon a collection of observations.
% These observations can come from simulation, nature, and other machine learning models.
% The concentration of machine learning can be defined as the process of addressing and solving any particular problems by collecting enough datasets and algorithm-based training a statistical approach to the collected dataset.
Machine learning approaches can be classified according to 
different criteria, here we  just mention the basic
paradigms of supervised, unsupervised and reinforcement learning, see an illustration in Figure \ref{fig:ml-subfields}.  The review will mainly deal with supervised learning and 
reinforcement learning under a perspective of their applicability in biochemical engineering.
%In this review, we do not intend to cover all theories of these mentioned subdivisions but rather concentrate on the most applicably principled explanations, mainly supervised learning and reinforcement, to the problem in biochemical engineering. 
Interested readers can refer to many complementary references on machine learning domains \cite{hutter2019automated,mohammed2016machine,murphy2022probabilistic,sra2012optimization}.

The general principles of machine learning are best explained for supervised learning, that is in fact regression and classification. 
%Let's start by denoting a vector $\textbf{x} \in \mathbb{R}^D$ and a scalar $y \in \mathbb{R}$ as one observation/sample/feature vector and its label/target respectively. 
A data source would produce pairs of inputs $x\in\mathcal{X}$ and outputs $y\in \mathcal{Y}$ coming from some unknown distribution
on $\mathcal{X}\times\mathcal{Y}$.  Usually $x$ is a vector of features and $y$ a vector or real numbers (regression) or class labels (classification).

The data obtained until some time 
would be collected in a dataset $D=\{ (\textbf{x}_1, y_1), (\textbf{x}_2, y_2), \dots, (\textbf{x}_N, y_N) \}$.
The goal of supervised learning is  to learn a predictive model from such a
dataset, namely a map $f:\mathcal{X}\to\mathcal{Y}, x\mapsto f(x)$ that predicts or estimates $y$ given $x$.  Such maps usually are obtained by specifying the parameters of a parametrized family of 
maps $f_\theta:\mathcal{X}\to\mathcal{Y}$ for $\theta$ 
in some parameter space. 
The quality of a prediction $\hat{y}$ 
is measured by a loss function  $\ell(y, \hat{y})$. 

Models are usually trained on  a dataset by minimizing
the so called empirical risk, which is the average loss
on a given 'training' dataset\
\begin{equation}
	\label{eq:empirical_risk}
	\mathcal{L}(\theta) \eqdef \frac{1}{N} \sum_{i= 1}^{N} \ell (y_i, f_{\bm{\theta}}(\textbf{x}_i)) \; .
\end{equation}     

The  minimization problem to be solved is
\begin{equation}
	\label{eq:theta_hat} 
	\hat{\bm{\theta}} = \argminA_{\bm{\theta}} \mathcal{L} (\bm{\theta}) = \argminA_{\bm{\theta}} \frac{1}{N} \sum_{i= 1}^{N} \ell \left(y_i, f_{\bm{\theta}}(\textbf{x}_i )\right) \; .
\end{equation}

With a quadratic loss function (the negative log likelihood of Gaussian noise on the $y$) this is the classical 
least squares fitting of the parameters of a parametric
regression model. %

The purpose of a predictive model is of course to give reliable
predictions for \emph{new} inputs, that is 
to generalize well. The best model in  the family would be the model with lowest \emph{expected prediction
error} (EPE), that is the lowest average loss over all possible 
data pairs 
$\mathbb{E}_{x,y}\ell(y,f_\theta(x))$ in the limit
for infinite sample size.

However, the empirical risk for the optimized parameters
\begin{equation}
	\label{eq:empirical_risk_opt}
	\mathcal{L} \eqdef \frac{1}{N} \sum_{i= 1}^{N} \ell (y_i, f_{\hat{\bm{\theta}}}(\textbf{x}_i ))
\end{equation}     
on the training dataset is often much smaller than the expected prediction error, namely
when the model is overfitting. 
A small empirical risk on a training set (or a good fit) is no reason to rely on a predictive model.

Overfitting can be an issue even in mechanistic models 
with only a few parameters  and lots of data, but it always
is in machine learning,  as an ML model would typically use 
a huge number of  parameters to be flexible 
enough for modelling an unknown relationship.  Therefore it is important to estimate the  EPE well, in order to be able to assess a model.

Cross-validation is the method of choice to obtain reliable estimates of the EPE  under fairly general conditions.  One divides the dataset   into a training dataset and a test dataset $D=D_{\mathrm{train}}\cup D_{\mathrm{test}}$.
The model is then trained on $D_{\mathrm{train}}$, while the 
average loss is reported on $D_{\mathrm{test}}$,
that is on samples never seen during training. Cross-validation somehow simulates applying the model to new data. That's the basic idea, although it is usually advisable to average this procedure over different splits (N-fold cross-validation); for  more sophisticated versions of cross-validation and best
practices of data partitioning see \cite{murphy2022probabilistic,mahmud2020survey,bussola2021ai}.

Apparently more
interdisciplinary communication would be necessary
to make the notion of model assessment by cross-validation well understood and accepted outside ML \cite{king_cross-validation_2021}.  There are some caveats: With small
amounts of available data cross-validation may be unfeasible.  Furthermore the structure of datasets 
can make  partitioning a complex task quite difficult. Different replicates of an experiment should for example all end up in the train or all in the test partition, otherwise
the test loss can underestimate the expected prediction error. Doing cross-validation right is of the essence and would usually require communication between 
a domain expert and a machine learning expert.

Machine learning models are constructed with different architectural choices and also with different regularization techniques to avoid overfitting, which leads to
a family of parametrized models instead of just one model. The family itself is parametrized by the so called hyperparameters that control architecture, regularization, training, etc.  Each model
defined by a specific set of hyperparameters has its trainable parameters to fit the available data.

Now we encounter the classical task of model discrimination and model selection in a new guise.  Finding optimal hyperparameters means
selecting the model in the family with the  best predictive performance,
that is the lowest expected prediction error.  
The estimation of the EPE
used in model selection should rely on a validation dataset $D_{\mathrm{validation}}$ different from the test set $D_{\mathrm{test}}$
used for the reporting the EPE of the selected model, otherwise
the latter may be grossly underestimated.   Thus e would need to partition 
the dataset in three disjoint datasets  $D_{\mathrm{train}}$, $D_{\mathrm{validation}}$ and $D_{\mathrm{test}}$.

Hyperparameter tuning \cite{feurer2019hyperparameter,yang2020hyperparameter} 
is an essential part of machine learning,  models
rarely work convincingly out of the box, which should not come as a surprise as even simple regularized regression methods like Ridge Regression and Lasso require tuning the regularization parameter in order to pay off.  Cross-validation comes with a high computational burden, which cannot be avoided.  But
it would not require additional mental effort of scientists
that want to apply ML, once frameworks automate  this
procedure.

% These values are not derived from the training data and can, therefore, not be learned by the model.
% For example, early\_stopping\_rounds is a hyperparameter in XGBoost \cite{chen2016xgboost,mitchell2017accelerating} model that indicates the number of rounds without improvements after which we should stop training.
% Hyperparameters can also derive from machine learning pipelines, such as the choice of imputation technique on the missing data.   

In machine learning, a baseline is any simple algorithm, with or without learnable parameters, for solving a task, usually based on a heuristic experience, randomization, or elementary summary statistics \cite{chen2020training,chung2021beyond}. 
This is an important reminder when tackling new domains with machine learning techniques, such as bioengineering and bioprocessing. 
Before attempting to develop more sophisticated models, obtaining existing simple baselines is more important. That is, Occam’s Razor or takes the simplet hypothesis which is consistent with the available data. All models and algorithms already established in the domain serve as baselines.    A sophisticated time series forecasting model for a bioreactor must be measured against existing reactor models, in order to assess it.

\subsubsection{When to Use Machine Learning?}
\label{sbusec:when-to-use-ml}

Machine learning is now extensively applied and is even a driving force of discovery all over science,
but it is no a panacea.
The prominent successes come at the price of the less prominent failures. 
Quite a few things can go wrong if not heeded,
leading to the risk of misinterpretations, over-optimistic results and models that fail to 
generalize.  Recommendations and best practices for the use of machine learning in science \cite{riley2019three}, 
or more specifically computational biology and biology  \cite{greener2022guide, chicco2017ten} can help
to avoid these mistakes and to save time and money.

%ubiquitous across all the fields that drive discovery in science.
%However, it is not a powerful tool for solving all practical problems.
%Many of the approaches are so complicated, or heavy-computing assumptions that it is impossible or expensive if they are applied in the wrong contexts  which might lead to enormous wasted scientific efforts, invested time and money but receive in high risk of misinterpretations, over-optimistic results, error-prone analyses and the illusion of successful conclusions.

When should we   consider deploying and investing in machine learning?

\paragraph{When it is cost-effective}  

It is difficult to know in advance when the application of machine learning will lead
to a cost reduction in bioprocess engineering.  Decisions for investments would ideally be based on
a comparison of the cost of alternatives \cite{wheelwright2020economic,ng2019bioprocess}.
Improved models and algorithms through machine learning may reduce the cost and experimental
burden of developing and scaling a bioprocess and possibly lead to more cost
efficient control of bioreactors at the industrial scale.  But applying machine 
learning would also create costs, namely for data management and curation, 
development of models, expensive hardware or cloud computing for training the models,
building  and running the infrastructure to deploy and monitor the complete machine learning project life cycle 
which includes further continuous monitoring the model, collecting new data and keeping the model up to date.
However, we believe that given the range of problems that might be solved by machine 
learning investment in such infrastructure seems reasonable.

% Improved models and algorithms for control and operation
% of bioreactors and improved methods for strain characeterization scale-up would save costs of bioprocess development
% and possibly render processes on the industrial scale more cost efficient.

% or improved methods for the characterization of strains would save
% costs of experimental facilities and operations needed for the development of a process
% \cite{wheelwright2020economic,ng2019bioprocess},
% but has to b
% The cost factors of bioprocess engineering derive from development, facilities, and operations  which drive fundamental 
% considerations for investment choices.
% From the machine learning perspective, the costs sum up from data collection, e.g., lab work, simulation data generation, preparing and cleaning data, training the models, e.g., the expensive hardware or cloud computing needed to train deep models, and building and running the infrastructure to deploy and monitor the complete machine learning project life cycle which includes further continuous monitoring the model, collecting new data and keeping the model up to date.

%\paragraph{When the problem has a simple objective}
\paragraph{When needing regression or classification with enough data  }

Whenever a problem in the domain can be formalized as one of the basic supervised  machine learning problems (regression, classification), and when there is a relatively large amount of  aggregated legacy data or the acquirement of new data fairly cheap,
machine learning can make useful contributions.  Of course,
it depends on the quality of the models already in use, whether
a significant improvement is to be expected. So it is above
all the experts knowledge of the deficiencies of the models
and algorithms they use which points to useful applications of
machine learning.

As mentioned above \ref{subsec:machine-learning} good data management and curation is an enabler for machine learning.  Data
with complete metadata annotations allow for
aggregation of data across different situations, e.g. bioprocess data for different strains, scales, reactors.   And
that is the situation where adequate
machine learning models are most advantageous.

% Domain problems have to be translated into machine learning. wh 

% When the model can be written down as a supervise
% As we mentioned in \ref{subsec:machine-learning}, target $y$ can be a real number, a class, or a vector at a particular time, depending on a practical problem. 
% We can convey and transform bioprocess engineering and formulate it as a machine learning problem with a concrete objective.
% Note that a machine learning model can only work if $y$ is consistent to a specific during training and inference afterward.
% For example, $y$ is a real number while deploying the model and training on a training set.
% It should also be a real number when we generalize it to a test set or other data source. 
% Obviously, we build as many models for bioprocess modeling and control problems.

\paragraph{When the data consists of images or videos}

For image classification, object detection, image segementation and other image related tasks machine learning models based on convolutional neural networks have consistently beaten all previous methods.
Automatic analysis of images allows to 
turn imaging devices into soft sensors. For example
microscopic images of samples from a bacterial fermentation give information on the heterogeneity of the population, inclusion bodies, shapes of bacteria. Building an automatic image analysis pipeline
for this purpose is fairly easy, using an open source library for
bacterial image analysis \cite{spahn_deepbacs_2022}. Manually annotated training data would still be 
mandatory.

\subsubsection{Machine Learning Project Life Cycle}
  
Machine learning is implemented as a process containing chained stages: Data cleaning techniques, data transformation or normalization, hyperparameter 
optimization using cross-validation, model training
and validation, deployment, monitoring and maintainance,
which includes updating trained models (and possibly also the hyperparameters) when new data come in or possibly
querying new data in order to improve the model (active learning)
\cite{ashmore2021assuring,kumeno2019sofware,luu2021managing,zaharia2018accelerating}. 
% We usually begin with the data, applying data cleaning techniques, transformation, training/validation/test division, then using machine learning models. 
% Next, we evaluate their performance, deployment, monitoring, and maintenance. 
% The hyperparameters of the entire process are often optimized based on existing data and model knowledge. 
% The whole process can be implemented and used for predictions 

When applying machine learning for bioprocess engineering
the specific problem has to be defined and then to be
formalized as a machine learning task or possibly
as a composition of several machine learning tasks
\cite{rathore2021bioprocess}.

If, for example, an application problem
problem can be framed as a supervised learning problem,
we have to specify which output quantity should be 
inferred from which input quantity, what is the relevant loss function to evaluate model predictions, what kind and the
amount of data that are available.  It might also be necessary
to specify requirements on train-test-splits.

%A bioprocess-based machine learning project begins with understanding a bioengineering problem that we may or may not tackle with machine learning. 

%It might also contain one or more different machine learning models if we can break down the bioengineering project into various sub-tasks  
%\cite{rathore2021bioprocess}.
% Once an engineering project is defined, we transform it into a specific machine learning problem, e.g., unsupervised, supervised, and reinforcement learning. 
% The result of applying machine learning should be materialized into a deterministic predictive model, i.e., what should be the input and output of the whole process? 
% Bioengineering experts can investigate a machine learning model as a black box. They do not care about algorithm design but the input data, output results, and hyperparameters, if any. 

After these specifications machine learning pursues a
well-defined goal. For supervised learning the 
procedure would try to obtain
a model with lowest expected prediction error among
all candidates.  If the loss function indeed reflects
the requirement of the engineers the model should be
useful for them.

% Note that the goal of machine learning is not necessarily the same as the goal of biological engineering. 
% For example, a machine learning m predicts final biomass before induction in a one-liter bioreactor setting and experimental configuration.
% We optimize a loss, e.g., root mean squared error, between the ground truth and predicted output where any proposed models outperform baselines indicated by a low standard deviation and high certainty.
However, the goal of a bioprocess engineer is more general, namely to achieve a technological objective with available resources. So the engineer
has to care about the cost  of lab work, monetary investment, and data collection  necessary for
a successful solution of the narrower machine learning
task.

\begin{figure}[ht!]
	\centering
	\includegraphics[width=1.0\textwidth]{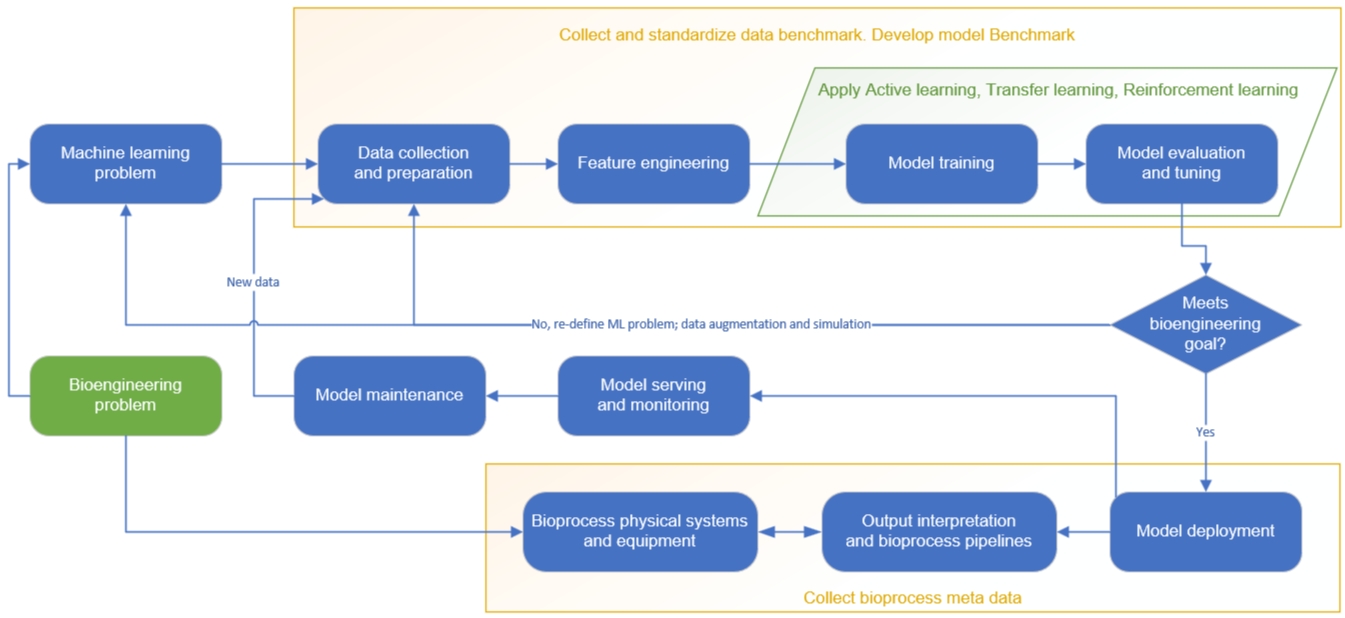}
	\caption{Machine learning bioengineering project life cycle.}
	\label{fig:ml-project-life-cycle}
\end{figure}

In general, the life cycle of a machine learning project, illustrated in Figure \ref{fig:ml-project-life-cycle}, consists of the following stages: 1) bioengineering, isolate a problem and rephrase it as equivalent equivalent machine learning tasks, 2) data engineering, e.g., data collection and preparation, feature engineering, 3) machine learning engineering, e.g., model training, model evaluation and tuning, model deployment, 4) machine learning in production, e.g., model serving, model monitoring, and maintenance \cite{habibi2014review,mey2021improving,panjwani2021application}.

%\subsection{Neural Ordinary Differential Equations}
%\input{tex/neuralode.tex}

\subsection{Active Learning}
\input{tex/active_learning.tex}

\subsection{Transfer Learning}
%In terms of ordinary perception, humans do not learn to perform tasks individually. 
Humans  do not learn to perform tasks independently, but always make use of previously acquired knowledge and skills, when dealing with a new task.
%our knowledge on the similar challenges we have learned before and adapt it to new ones. 
It is an important challenge for machine learning to
find ways that mimic this human connectivity of knowledge
and allow to reuse information from other contexts in a new one. Machine learning approaches 
that try to \emph{reuse} (parts of) models trained for one task in a new task are known as \emph{transfer learning}
%, namely how to reuse knowledge we already learn, particularly by parameter values, and start processing new data based on existing parameter values. 
%In machine learning methodology is called transfer learning 
\cite{torrey2010transfer, weiss2016survey,rostami2021transfer,long2017deep,Duong-Trung2019}. 
There are other notions to be distinguished from transfer learning, which also imply
'learning from other cases', namely \emph{meta learning} and \emph{multi-task learning}.
Meta learning applies, when a task refers to datasets drawn from a distribution
of datasets, for example with each dataset a collection of
bioprocess data for a specific strain of \textit{E.coli}. The goal is to devise models that will train 
faster on a new dataset, using information from other datasets. 
Multi-task learning, on the other hand,  is closely related to transfer learning, but
it deals with \emph{simultaneously} training models for different tasks instead of \emph{reusing} pre-trained models.  

Most applications of transfer learning refer to  learning tasks where the inputs $x$ have the same data type and can be assumed to be similar in some sense, e.g. images of a certain
format,  protein amino acid sequences,  
fermentation data with the same observables
in the same format. The outputs $y$, however, can be specific to the different tasks.

Reusing  models trained for different tasks can be a very 
cheap way to overcome the 'small data problem'.

% This way, machine learning models do not start learning from scratch for previously handled problems. 
% This has significant implications for bioprocessing techniques in that the data is incomplete for a single model training and requires the machine learning model to continue training itself when new data is available.
% It gives a considerable advantage over traditional approaches that require training from scratch, are computationally expensive, and demand large amounts of data per training to achieve high performance.
% Transfer learning, however, attempts to change it by developing methods to transfer knowledge learned in one or more source tasks and use it to improve understanding in a related target task \cite{christopher2018performance}.
Neural networks for image classification  trained
on very large image datasets
\cite{deng2009imagenet,russakovsky2015imagenet} have led to the arguably most successful
applications of transfer learning \cite{huh2016makes}. Such models process
the original image through subsequent stages, each stage
producing a new representations.   These representations 
capture image features, some very general and useful
outside the original training context, some very specific
to the original training task.   For a new task on a small
dataset, one can use a part of the trained network as a component
of a new model and then train the model on the new data. 
This has been successfully applied to many different image domains (medical histology, plant images, etc.)
\cite{christopher2018performance, duong2019classification, duong2019combination,  tran2020recognition, duong2021classification}.

A biochemical and biotechnological use case of 
transfer learning 
that is slowly unfolding its potential is
the prediction of protein properties
from the underlying sequences.  In the very large
protein libraries some protein properties are more
frequently available than others.  Most proteins 
are equipped with class labels in a protein classification, fractions of the proteins have
3d structures, enzymatic activities, physico-chemical
properties attached.  A model trained for predicting
some of the frequently available properties or for 
an unsupervised task on all available protein sequences
may learn  internal representations that are also useful for other tasks related to rarely available properties
\cite{alley_unified_2019,bepler_learning_2019,yang_learned_2018,rao_evaluating_2019,wittmann_machine_2020}.  A regression model on a low-dimensional
representation may be trainable with only a few examples, whereas any
model that directly works on the high-dimensional space of amino
acid sequences needs large amounts of data.  If protein properties prediction
models are good enough at ranking potential protein variants
they can speed up directed evolution \cite{wittmann_machine_2020}.

Transfer learning has also been applied to modelling lutein production
by microalgae \cite{rogers2022}. For one species of microalga a comparatively large amount of published data was available, for a second
microalga with a somewhat similar growth behaviour there were fewer data.
Models were then trained on the data for one species and then transferred
to models for the second species.  Some data augmentation was done in both cases to improve training.

\begin{figure}[ht!]
	\centering
	\includegraphics[width=0.9\textwidth]{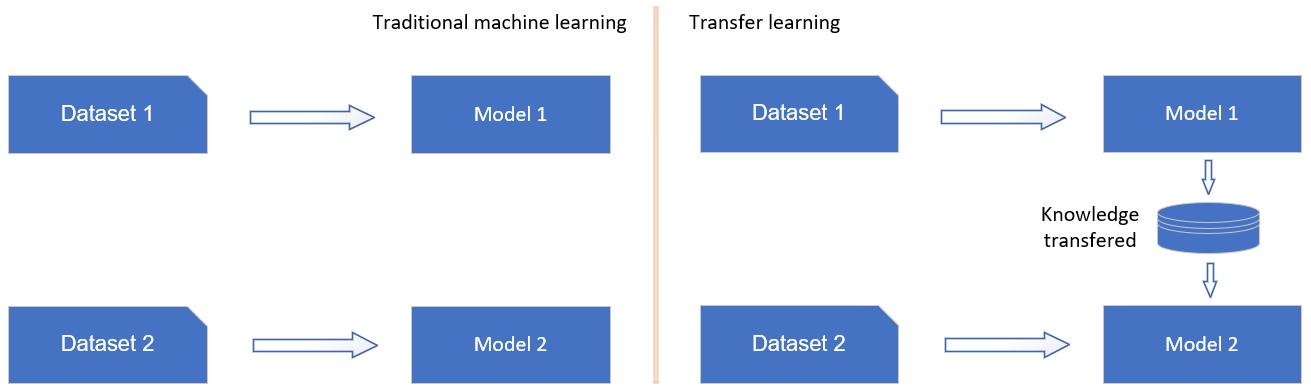}
	\caption{Difference between traditional machine learning and transfer learning.}
	\label{fig:transfer-learning-vs-traditional-machine-learning}
\end{figure}

The particularities of transfer learning are presented in Figure (\ref{fig:transfer-learning-vs-traditional-machine-learning}).

\begin{figure}[ht!]
	\centering
	\includegraphics[width=0.5\textwidth]{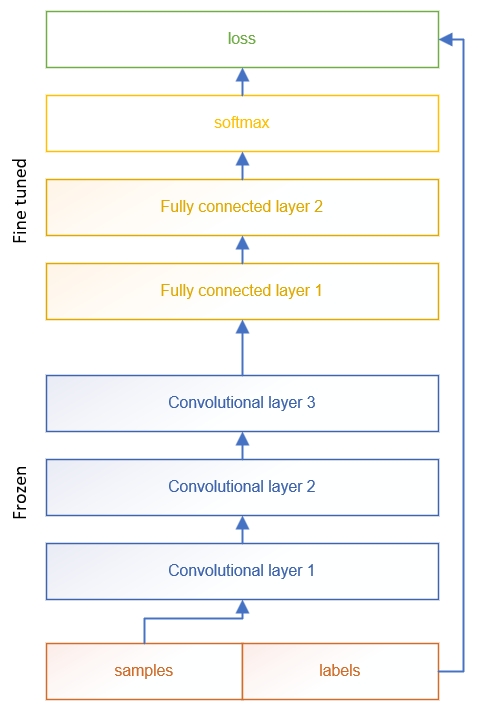}
	\caption{Frozen (no update during training) and fine-tuned (update during training) layers.}
	\label{fig:freeze-fine-tune-layers}
\end{figure}

\begin{figure}[ht!]
	\centering
	\includegraphics[width=1.0\textwidth]{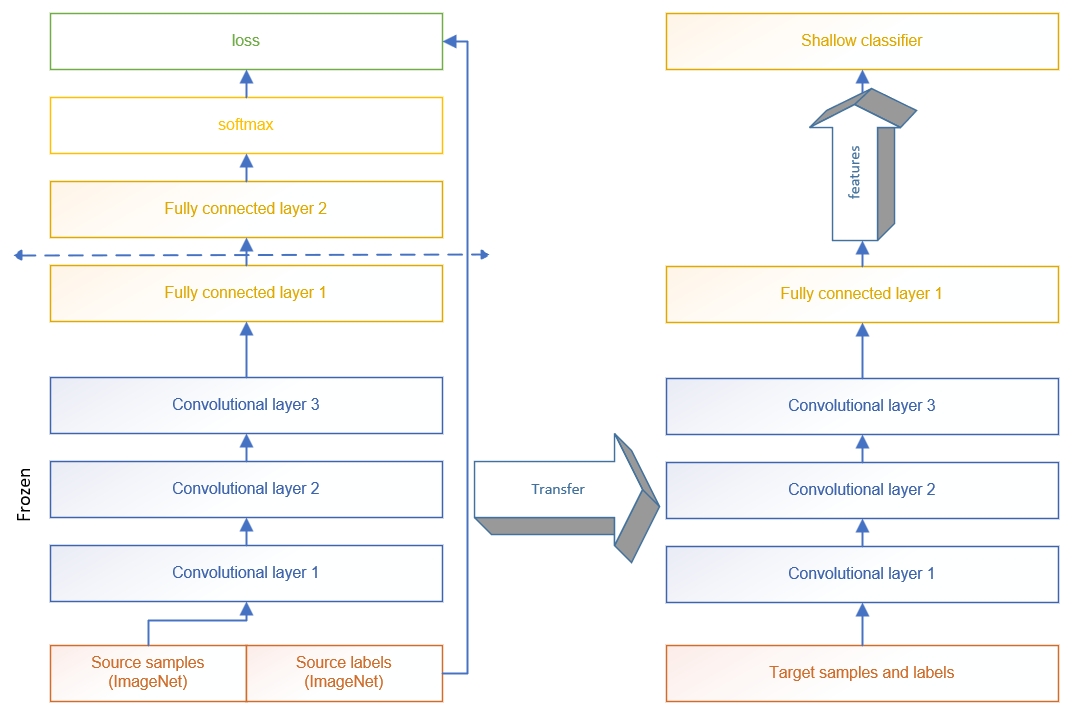}
	\caption{Initial layers as feature extractors.}
	\label{fig:feature-extractor}
\end{figure}

When training  a model with a pre-trained 
part one can decide which of the  inherited
parameters will be frozen and which will be retrained
with the new model. Thus the pre-trained model
either serves as
a feature extractor \cite{mahajan2021plant,izadpanahkakhk2018deep}, see Figure \ref{fig:feature-extractor}, or as an initializer \cite{neyshabur2020being}, see Figure \ref{fig:freeze-fine-tune-layers}.

 In transfer learning learning rate and number of training epochs  
 correspond to a trade-off between the influence of the data from the original domain and the influence of the new data
\cite{duong2020towards}.  The optimal amount of training and the
best architecture for the task have, as always, to be determined
by cross-validation.

\subsection{Reinforcement Learning}
\label{sec:reinforcement_learning}

Reinforcement learning (RL) is one of the main branches of machine learning. While the supervised and unsupervised learning methods learn the model from a given data set, RL methods learn to act. In other words, the outcome of the RL is the optimal decision rule for a given state, which is also referred to as `policy' given an objective \cite{Sutton2018}. In general, RL performs the following three-step procedure iteratively: data generation, performance evaluation, and policy improvement. 
By interacting with the dynamic systems according to the policy, the RL agent receives the data consisting of states, actions, and rewards. The data is used as a reinforcement signal that evaluates the performance of the policy. The policy is improved based on the performance evaluation with various types of optimization methods. The procedure is usually designed to be stochastic, not only to act against the uncertain systems but also to add exploratory actions to the system to prevent trainable machine learning models from being overfitted \cite{yoo2021reinforcement}. This addresses the trade-off between exploration and exploration explicitly. 

RL is deeply connected with process control in the sense that RL solves sequential decision-making problems \cite{Bertsekas2005dynamic}. RL has several potential advantages over the standard approaches of bioprocess control that use mechanistic models and mathematical programming. First, RL is flexible to work with varying levels of mechanical knowledge and structure of the systems \cite{busoniu2017reinforcement, kim2020gdhp, kim2020convergence}. Model-free is a special characteristic that distinguishes RL from other process control methods, and the reinforcement signal is solely used for policy improvement. Therefore, model-free RL can handle (1) hybrid systems consisting of mixed continuous and discrete states, actions, and events, (2) problems with various objective functions encompassing tracking control, economic optimization, and experimental design, and (3) model uncertainties that are not restricted to Gaussian distribution. This flexibility is an appealing characteristic for the bioprocess control and optimization \cite{lee2005approximate}, because biological models are often challenging to build, and biological systems have a considerable level of uncertainties. Recent advances in statistical machine learning enable feature analysis of the raw sensory-level data by using deep neural networks and the implementation of various information-theoretic techniques. Synthesis with a deep learning framework, deep RL (DRL) has successfully achieved a scale-up of RL methods to high-dimensional problems, showing remarkable performances in various applications such as process scheduling, reaction mechanism, fluid dynamics, robotics, autonomous driving, etc. \cite{kim2018pomdp, oh2021automatic, horwood2020molecular, novati2021automating,Levine2016, williams2016aggressive, Silver2017}.

The RL's second advantage is that most of the computation is done off-line. In contrast, the conventional mathematical programming approaches need consistent re-planning, which can lead to exorbitant on-line computational demand. Because a single model cannot perfectly characterize the complexity of the metabolism, bioprocesses have to be operated in a closed-loop manner, adapting the model to the most recent experimental data \cite{lucia2017rapid}  However, the mathematical programming-based approaches such as model predictive control (MPC) cannot match the online computation limit when the complexity is high due to the combination of the model, operating constraints, and uncertainties. Several researchers have focused on the RL framework as a complementary method \cite{bucsoniu2018reinforcement, lee2006choice, lee2009approximate}. It is the nature of RL that the policy is obtained, essentially the closed-loop feedback rule concerning states of the system. An end-to-end closed-loop operation can be achieved if the RL is applied to industrial bioprocesses using massive historical raw data in an offline environment.

Motivated by these advantages, several pioneering pieces of work for the bioprocess control were first appeared in \cite{wilson1997neuro, peroni2005optimal, li2011reinforcement}. In these studies, the RL methods use the lookup table that measures the optimality (e.g., `cost-to-go' function or Q-function) with respect to the discretized the state and action space. \cite{wilson1997neuro} used a fuzzy lookup-table guided by the expert knowledge in the frame of a Q-learning algorithm. It showed that the RL could achieve near-optimal performance for the batch process control. \cite{li2011reinforcement} combined the fuzzy rule with the Q-learning method to determine the gains of a PID controller for a tracking problem of the fed-batch bioreactor. \cite{peroni2005optimal} solved a free-end problem for a fed-batch bioreactor using approximate dynamic programming, an analogous algorithm to the RL. The RL algorithm was tested under different initial conditions and showed optimal performance without additional recomputation. This is the first work that recognizes the merit of RL for the closed-loop operation in the presence of disturbances. 

Recent works incorporate DRL methods, which allow for an extension to the optimization under the continuous state and action space \cite{pandian2018control, ma2020machine, petsagkourakis2020reinforcement}. In \cite{pandian2018control}, partially supervised RL was used to solve a tracking problem of a yeast fermentation problem. Neural networks that map the state and setpoint with the control input were trained and refined using RL. A DRL algorithm, asynchronous advantage actor-critic (A3C), was incorporated into the biomass maximization problem of a fed-batch bioreactor \cite{ma2020machine}. 
\cite{petsagkourakis2020reinforcement} utilized the policy gradient method for the optimization and recurrent neural networks (RNN) to approximate the policy function. The RL method was performed preliminary using offline data, and the policy was further trained in the online implementation. 

The main drawback of model-fee RL is that it is notoriously difficult to use due to the sensitivity to hyperparameters, intractable data requirement, and optimistic estimation of the Q-function values \cite{henderson2017deep, fujimoto2018addressing, kim2020gdhp}. Even for the optimal control of the most straightforward linear system with the quadratic objective function, model-free RL fails to achieve a reliable solution compared to the standard linear quadratic regulator algorithm \cite{recht2019tour}. This limits the actual applications to the control and optimization of the real bioprocesses. Model-based RL can help solving the issue by using the mechanistic model as a simulator for the offline training, or utilizing the model equations' gradients to accelerate the training \cite{langlois2019benchmarking}. \cite{kim2021model} suggested a two-stage optimal control for a closed-loop dynamic optimization of a fed-batch bioreactor. In the high-level optimizer, differential dynamic programming, a model-based RL that uses model gradient, is used for the long-term planning with the economic objective of maximizing productivity. Whereas in the low-level controller, MPC is used for the short-term planning that tracks the high-level plan and, at the same time, rejects disturbances and model-plant mismatch. \cite{oh2022integration} proposed the integrated formulation of the MPC and RL, where the terminal cost function of the MPC is replaced by the value function obtained by the model-free method. The method was validated for the optimization of an industrial-scale penicillin bioreactor.

Another issue about RL is the consideration of critical process constraints for safety and keeping the operating condition within the valid domain \cite{yoo2021reinforcement}. A typical way to consider process constraints is to augment the amount of constraint violation as the penalization term to the objective. Using augmentation solely cannot always guarantee the feasibility of the exploration. In \cite{pan2021constrained, petsagkourakis2022chance}, the probability of constraint violation was formulated as chance constraints, and an adaptive back-off approach was implemented to reduce the violation. Nevertheless, the trade-off between the original objective and constraint penalization is not uniquely determined, therefore adding another hyperparameter to the overall algorithm. This is not the case in conventional mathematical programming-based approaches such as MPC. In this regard, model-based RL can be useful. \cite{mowbray2022safe} suggested Gaussian processes regression for the data-driven state-space model and model-based RL for fed-batch fermentation processes. Mechanistic model-based RL approaches \cite{kim2021model, oh2022integration} can naturally address constraints of fed-batch bioprocesses.

%% file: tex/active_learning.tex
\subsubsection{What is Active Learning?}

ML models are usually trained on large corpora of data created by a potentially unknown process.
As stated in Section \ref{subsec:machine-learning}, supervised machine learning solves a regression or classification problem that requires the data to be given  or to be representable as predictors $x$ and target values or labels $y$. 
In some contexts, for example image classification, the target values are also known as annotations. The term refers to a situation where a large data set of predictor data $x_i$ is available or continuously generated, but human domain experts are needed to annotate, i.e. provide the  label
$y_i$ for some of the data points $x_i$, an expensive and time-consuming procedure. 
In other contexts the acquisition of new data  is inherently expensive,
for example if the data are obtained by chemical, biochemical or biological
experiments or complex computer simulations. In all cases the cost of data acquisition  and
a limited budget force ML practitioners to select which data should be acquired or annotated 
to be most informative for the model. This process is called Active Learning (AL).

% However, when new data are produced, models can improve if they can actively choose which data to be acquired. 
%
% [Thorben] I am currently omitting the statement about increased performance until I find a good reference for such a case

The active learning task to query new data beneficially can be seen as a generalization of the classical problem of (sequential) optimal experimental design (OED).  
Experimental designs can be chosen optimally for different purposes, e.g., to discriminate model hypotheses, estimate model parameters, or predict at specific points.  

\begin{figure}[ht]
	\centering
	\includegraphics[width=0.9\linewidth]{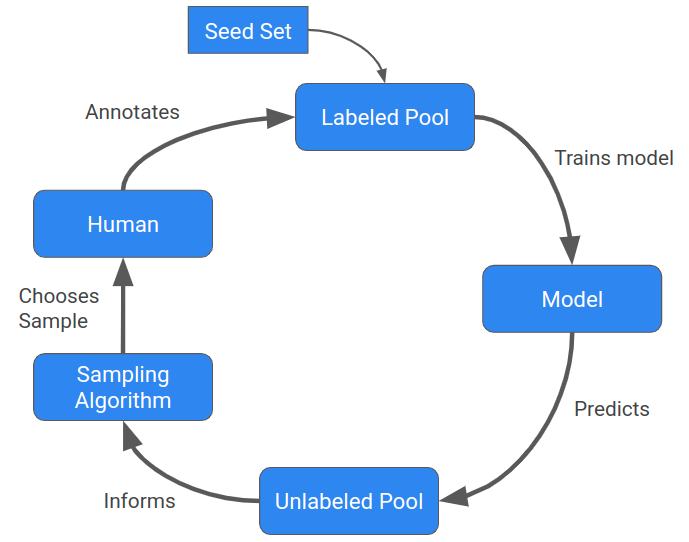}
	\caption{The active learning cycle. The seed set contains a small number of labelled samples. Each cycle adds one or more datapoints to the labelled pool. Repeated until some stopping criterion is met (usually a budget constraint or performance threshold).}
	\label{fig:al_cycle}
\end{figure}

Since active learning methods are incremental (selecting the next data point based on the current labelled data and model), they often require a so-called seed set.
This is a small set of labelled data $(x_i,y_i)_{i\in I_l}$ used to train the initial model and calibrate the AL method.
For a graphical overview of the AL cycle see Figure \ref{fig:al_cycle}.
The labelled set is initially comprised of the seed set. Each cycle adds one or more datapoints to the labelled pool.

\subsubsection{Different Sampling Scenarios}
%Let us consider data consisting of independent variables $x_i\in\mathcal{X}$ and additional target variables $y_i\in\mathcal{Y}$ with some unknown joint distribution of pairs $(x_i,y_i)\in\mathcal{X}\times\mathcal{Y}$. \\
%Active learning is generally data and model agnostic. It has been applied to many different combinations of data modalities and model types. 
One distinguishes three scenarios as to the way the next data point is sampled.
%Different methods are categorized by the sampling algorithm that select the next data point. \\
%Active learning distinguishes three scenarios:

\begin{description}
	\item[Stream-based Selective Sampling] Data $x_i$ is presented in a stream, e.g., images arriving from a camera, and a cost is incurred for acquiring target values $y_i$, e.g., labels by a human expert. 
	The active learning algorithm has to decide on a case-to-case basis if a sample is to be labelled or not.
	\item[Pool-based Sampling] A large pool (or a subset thereof) of unlabelled instances $(x_i)_{i\in I_u}$ is given.
	The AL algorithm has to pick one or more data points from the pool, which are to be labelled $y_i$. 
	\item[Query Synthesis] The AL algorithm uses the current labelled data $(x_i,y_i)_I$ to synthesize new cases $x_i$ for which the target value should be queried.
	This does not rely on existing unlabelled data as in the previous two scenarios but creates
	new data. These data might e.g. correspond to a real world experiment described by  parameters 
	$x_i$, the  outcome of which becomes the target value $y_i$. 
\end{description}
For all three sampling scenarios, there are potential applications in chemical engineering and bio-engineering.   For tasks like  anomaly detection in processes one would have a pool of legacy
data with a partial annotation of anomaly, an incoming stream of
new data without annotation and would chose the cases, for which
to require an expert annotation \cite{rychener_architecture_2020}. 
However, the query synthesis case,
is arguably the most important in the biotechnological context: New experiments 
are designed in order to produce the most informative data.

One should be aware that the distribution of the queries can be expected 
to differ from the distribution of the ordinary data generating process. 
If, for example,  AL/OED is used to optimize a bioreactor model
for later use in the control of the reactor,  it is entirely possible that the regular operating regime is different from the data distribution that creates a strong predictive model.

%One should be aware that the distribution of the queries can be expected to differ from the distribution of the ordinary data-generating process.
%If, for example,  AL/OED is used to optimize a bioreactor model
%for later use in the reactor's control, it is entirely possible that the regular operating regime is different from the data distribution that creates a strong predictive model.
%The assumption is that the created dataset, might it be skewed or not, contains all relevant use cases of the problem resulting in a model that generalizes nonetheless.% Due to the significant model-process mismatch in biological systems is important that any data gathering bias is explicit and related to the objective or purpose for which the model is created%

\subsubsection{Querying the Most Informative Data}

A good intuition is that AL and OED will query the most 
informative data for the purpose at hand, although not all algorithms define ''most informative" in the same way. 

The  expected information gain (EIG), 
which is the expected reduction of entropy by the queries,
is an ideal Bayesian  utility function for AL to optimize, however estimating
the EIG is computationally very challenging. Most of the AL
methods below use a different objective, but a unified 
view (\cite{settles_active_2012}) is possible, which
explains their relation to the EIG.  
All the following methods use different proxies of EIG to select 'informative' queries.

\begin{description}
	\item[Uncertainty Sampling] 
	Beginning with \cite{lewis_sequential_1994} this is a widely used class of methods with different underlying estimators of uncertainty. While working well in some 
	instances, such algorithms can over-sample regions of the space $\mathcal{X}$ where noise dominates. \\
	The most prevalent measurement of uncertainty is the Shannon entropy applied to the output of a classification model (see Fig. \ref{fig:al_uncertainty} (c)). 
	If the model assigns a high probability to one class and low probabilities to all others, the entropy is between those probabilities is low.
	If the model assigns an equal probability to all classes (the model is uncertain) the entropy is high.
	\begin{figure}[h]
		\centering
		\includegraphics[width=0.95\linewidth]{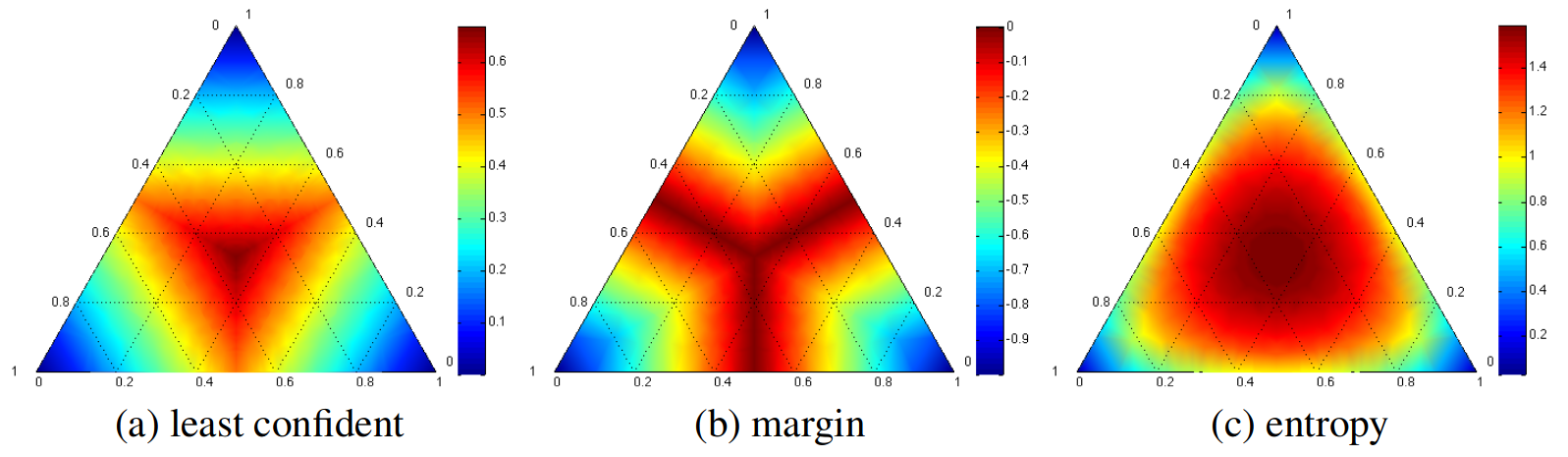}
		\caption{Different uncertainty measures applied to a 3-way classification problem (Ref. \cite{settles_active_nodate} Fig. 5). Each corner represents one class. Each point within the triangle represents the prediction of a model given an arbitrary datapoint. Each point indicates the assigned probability to each class by its position. The colour indicates the amount of estimated uncertainty, where red and blue indicates high and low uncertainty respectively.}
		\label{fig:al_uncertainty}
	\end{figure}
	A sample is considered informative if it produces high entropy in the model's output.
	\item[Reducing the version space] 
	Several approaches can be described as reducing the
	space of hypotheses compatible with the data, the so-called version space.% Due to their simplicity, bioprocess modeling requires choosing among alterantive kinetic models%
	These approaches maintain an ensemble of many models rather than just one.
	Each model represents one hypothesis about the available data (see Fig. \ref{fig:al_version_space}).
	An informative sample is considered one that produces high disagreement between the hypotheses/models, forcing wrong models to be dropped or updated.
	Repeating this process will push all models of the ensemble to converge to the ''true" hypothesis.% This hypothesis hevealy depends on the objetive of the modelling cycle%
	Algorithms that fall in this class are Query by committee and Query by disagreement.
	This method was implemented by \cite{beluch_power_2018} and applied to image classification tasks.
	\begin{figure}[h]
		\centering
		\includegraphics[width=0.95\linewidth]{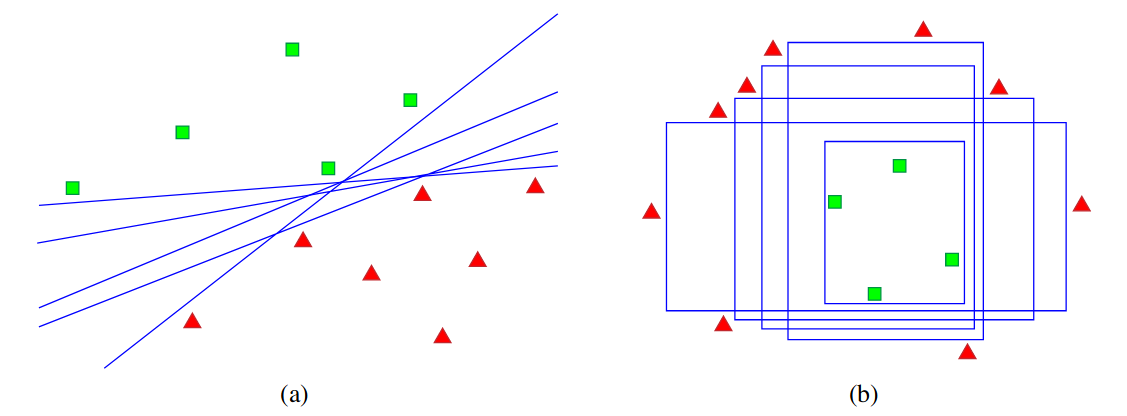}
		\caption{Different hypotheses of classification models for two types of classifiers (Ref. \cite{settles_active_nodate} Fig. 6). Each line or box respectively represents one correct hypothesis about the given data. An informative sample would be any new point that contradicts one or more of these hypotheses.}
		\label{fig:al_version_space}
	\end{figure}
	\item[Expected error reduction] 
	Another proxy of EIG is an estimate of the expected error after seeing a new query.
	Since all ML approaches rely on some error function to optimize their models, one can estimate the expected improvement in these functions given a selected data point from the unlabeled set.
	Points associated with a more considerable error reduction are considered more informative.
	This method has been successfully applied in classification (\cite{Roy2001TowardOA}).
	Since estimating the error reduction is computationally very expensive, \cite{konyushkova_learning_nodate} trained a regression model to predict the error reduction and applied it to 3D Electron Microscopy (Striatum) and MRI brain scans (BRATS).
	%
	% [Thorben] I'll leave the last two untouched. Stefan seems to know more about this than me
	\item[Variance reduction] That is where the classical 'alphabetical' frequentist OED criteria can 
	be placed. In maximum likelihood estimation, the covariance of the parameter estimates
	is bounded below by the inverse of the Fisher information matrix (Cramér-Rao). Different 
	criteria that control the eigenvalues of the Fisher information are used to find an
	optimal design (D-optimal determinant, A-optimal trace, etc.). Controlling the variance of the 
	parameter estimates has an impact on the predictive variance of the model but 
	is still a different problem. A-optimal design, however,  implies minimizing
	a lower bound on the predictive variance. Calculating and inverting the Fisher
	information matrics for all parameters of a large ANN is prohibitive, but recently
	this approach has been applied to just the last layer of an ANN \cite{ash_gone_nodate}.
	It should be noted the  Crámer-Rao lower bound may underestimate the true variance, and that the variance
	can be a poor descriptor for non-Gaussian distributions. 
	\item[Minimizing the EIG] 
	A direct estimation of the EIG has usually been considered an intractable problem, but recently useful (sharp) upper and variational lower bounds have been discovered and exploited for Bayesian Optimal Experimental Design (BOED) (\cite{foster_deep_nodate},\cite{foster_unied_nodate},\cite{foster_variational_nodate},
	\cite{kleinegesse_bayesian_2020}, \cite{kleinegesse_efficient_2019}, \cite{ivanova_implicit_2021}). These promising
 approaches still remain to be tested in the 
 context of bioprocess engineering. 
\end{description}

\subsubsection{Learning How to Active(ly) Learn}
There are three issues to raise with the previously mentioned methods.\\
(i) Most design criteria, even the theoretically sound ones, do not directly improve the utility
of the predictions for the final purpose.  \\
Increasing a model's information content or generalization capabilities 
is excellent, but the exact relation to a specific prediction task
or decision problem is not apparent. Therefore it would be advisable to \textit{learn} an
active learning strategy from data for the final task, end-to-end. 
It directly connects a model's performance on a given task to the selected queries. \\
(ii) If a new set of queries or experimental designs are selected each time, a complex nonlinear optimization problem has to be solved. This might require more time.
In a real-time setting where, i.e., experiments have to be redesigned based on new incoming data, the sampling process also needs to be fast %to account for the state dynamics of interest. \\
(iii) All methods discussed so far rely on heuristics to select their samples. 
These heuristics only use a limited subset of the available information and do so in a static fashion that does not adapt to the presented data. \\[2mm]
All three issues can be addressed when recasting the problem as reinforcement learning (RL) by parameterizing a policy that selects %or generates% the samples.   
This policy is learned %off-line% based on trial-and-error and a relevant reward function that defines the value selected samples in a sequence regarding model performance.
Crucially, the on-line application of the policy to %generate$ new data is straightforward (solving (ii)). 
In this setting, the policy is trained end-to-end concerning the final use of the prediction model, thus avoiding a possible mismatch between the optimization objective and the final use case (solving (i)).
Finally, the policy is usually represented by an ANN, so it can incorporate large quantities of information and data and dynamically learn how to utilize them (solving (iii)).
This includes summary statistics about the unlabeled pool (\cite{vu_learning_nodate}) or additional information about the model prediction and confidence (\cite{fang_learning_2017}).

This makes 'learning to actively learn' one the most promising approaches 
for AL \cite{fang_learning_2017, liu_learning_2018, vu_learning_nodate}, \cite{bachman_learning_nodate}.
Some of the recent BOED methods mentioned above (\cite{foster_deep_nodate}, \cite{ivanova_implicit_2021})
are policy based, too. 

If no sufficient legacy records of selected samples and improvement in model performance are available, one needs to employ more complex reinforcement learning approaches.
The authors of \cite{vu_learning_nodate} use model-based RL to solve the problem.
The agent is primarily trained within a simulation of the AL process and further improved based on the limited real-world data.

The interested reader can refer to section \ref{sec:reinforcement_learning} or directly to \cite{Sutton2018} for an introduction into reinforcement learning.

\subsubsection{A Special Case}

All previously described AL methods are done by analytical and probabilistic models.
However, there are also discrete problems amenable to logical analysis in the application domain of this article.
The most prominent example is the Robot Scientist Ada \cite{king_functional_2004},
an automatic system that designs experiments to determine the gene function of yeast
using deletion mutations and auxotrophic growth experiments.
The active learning strategy of Adam can  be formally understood along
the previously sketched lines
as reducing the hypothesis space by minimizing a probabilistic 
objective function (expected cost \cite{bryant_combining_nodate}). 
However, the gene network to be deciphered is 
treated as a logical problem, and a central ingredient of the algorithm is automatic logical
reasoning. This is an interesting case that recalls the ambiguous meaning of 
'artificial intelligence, which can refer to logical reasoning systems and
to statistical learning models alike.

In real-world scenarios, logical reasoning about complex, encoded information and
statistical learning on collected data can both play a role, though the
great successes of machine learning of the latter kind have recently eclipsed
the former.

\subsubsection{Spotlight: Uncertainty Quantification}

% [Stefan]:
% can you provide some more references and explanations
% I already put some references in the bibliography without
% referencing them
%
% [Thorben 18.05]:
% Viele der Infos von diesem Kapitel habe ich bereits in den vorherigen Text absorbiert.
% Ich werde versuchen, dieses Kapitel in ein Spotlight umzuwandeln, um Uncertainty Sampling nochmal genauer zu erklären, da das die meistgenutzte Technik für AL ist

\begin{figure}[h]
	\vspace{8mm}
	\centering
	\begin{subfigure}{.49\textwidth}
		\centering
		\vspace{-6mm}
		\includegraphics[width=.8\linewidth]{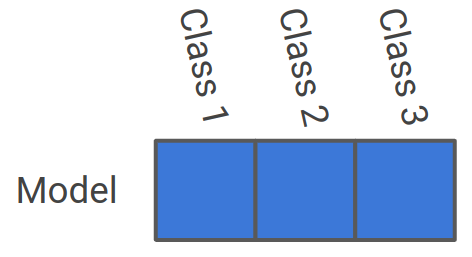}
		\caption{Model output}
	\end{subfigure}
	\begin{subfigure}{.49\textwidth}
		\centering
		\includegraphics[width=.8\linewidth]{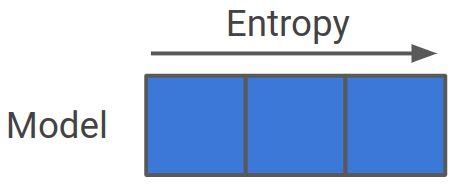}
		\caption{Classic uncertainty sampling}
	\end{subfigure}
	\begin{subfigure}{.49\textwidth}
		\centering
		\includegraphics[width=.8\linewidth]{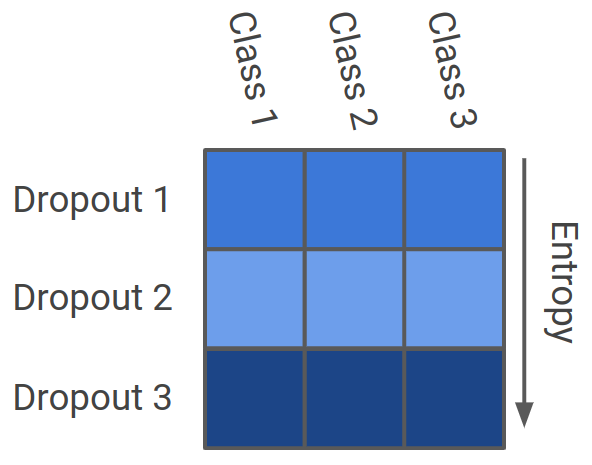}
		\caption{Uncertainty sampling with MC Dropout}
	\end{subfigure}
	\begin{subfigure}{.49\textwidth}
		\centering
		\includegraphics[width=.8\linewidth]{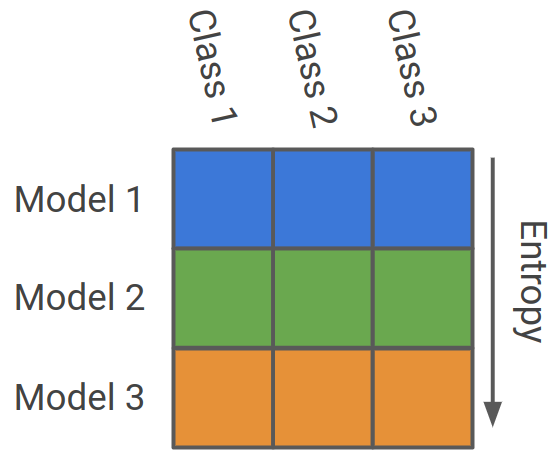}
		\caption{Classic uncertainty with ensembles}
	\end{subfigure}
	\caption{Comparison of different setups for uncertainty sampling with entropy for a 3-way classification problem}
	\label{fig:uncertainty_comparison}
\end{figure}

This section aims to deepen our understanding of uncertainty sampling, as it is the most straightforward and most used implementation of active learning. \\
As stated above, uncertainty sampling aims to measure the model's confidence for a given prediction and uses this as a proxy for the EIG.
The more uncertain a model is, the more informative this sample is considered, and following that, the more useful this sample will be when meaningfully annotated.
Figure \ref{fig:uncertainty_comparison} compares different setups for uncertainty sampling with entropy as uncertainty measure.
We will consider a 3-way classification problem so that the model will assign one probability for each class (subfigure (a)).
The classic uncertainty sampling will compute the entropy across the three classes (subfigure (b)).
Query-by-Committee was previously introduced as an alternative to uncertainty sampling since it was derived from a different theoretical motivation. However, the measurement of uncertainty in both frameworks is very similar.
Since Query-by-Committee algorithms maintain an ensemble of many models, uncertainty can be measured on a per-class basis (across models) rather than per model (subfigure (d)).
To assign a scalar value to each sample, the per-class uncertainties are usually summed (\cite{beluch_power_2018}).
Since maintaining and updating many ANNs is computationally very expensive, some methods try to simulate an ensemble by using an approach called Monte-Carlo Dropout (MC Dropout) (\cite{gal_uncertainty_2016}).
For MC Dropout, only a single ANN with Dropout-Layers is trained.
During prediction, where a single forward pass with a dropout rate of 0 is usually done, MC Dropout performs multiple forward passes with non-zero dropout, resulting in slightly different versions of the prediction.
Treating each forward pass as a separate model in an ensemble, the same Query-by-Committee algorithm can be applied (\cite{tsymbalov_dropout-based_2018}, \cite{beluch_power_2018}) (subfigure (c)).

% MISSING:
% Core-Set

%Most of the AL/OED methods rely on an uncertainty quantification of the model.  
%Predictive models used may be explicitly probabilistic (Bayesian Neural Networks, Gaussian Processes). 
%However many successful ML models are trained and evaluated for point predictions, so that other means to provide uncertainty quantification for such models
%become relevant. A rather general procedure consists in fitting ensembles of models (using e.g. randomized initializations, bootstrapping) and use the
%ensemble's predictions for uncertainty quantification. 
%For expensively trained deep neural networks, which represent a particularly successful model class, an a-posterio uncertainty quantification of single fitted model would be very useful.   It has been shown \cite{gal_uncertainty_2016} that for models trained with dropout using dropout at prediction can provide estimates of the predicted probability distribution.
%This method has been exploited  for active learning \cite{tsymbalov_dropout-based_2018}, but it may underestimate
%the uncertainty of the model far from the training examples. Surrogate
%models trained on the uncertainty estimates on the training set
%can be used to enforce the extrapolation property to become
%less confident with increasing distance to the training set 
%(\cite{tsymbalov_deeper_2019} with a Gaussian Process as surrogate). 
%For any such method additional care has to been taken that the
%probabilistic description is well-calibrated.  A naive belief
%in a probabilistic model may mislead the active learning.

%% file: tex/State-of-the-art.tex
%\section{Current State-of-the-art of bioMoCoML}

%Typical machine leanring solutions to address bioprocess modeling and control combine three key components: datasets, models, and optimization.
%Each components can interact with prior (single or multiple) knowledge of biochemical engineering. 
Machine learning has significantly contributed to the development of bioprocess engineering, but its application is still limited, hampering the enormous potential for bioprocess automation.
In this section, we summarize recent research across several important subfields of bioprocess systems, see Table \ref{tab:app-ml-bio-eng}, including bioreactor engineering \cite{xiao2019current}, biodevices and biosensors \cite{sode2016biocapacitor,dai2018single,pradhan2020nature,mehrotra2016biosensors}, biomaterials engineering \cite{ong2014introduction,tanzi2019foundations,dos2017engineering}, and metabolic engineering \cite{lawson2021machine,woolston2013metabolic,chae2017recent,presnell2019systems}. 
Bioreactor engineering studies the correlation and effects between complex intrinsic factors that operate a bioreactor (e.g., contaminant concentrations, temperature, pH level, substrates, stirring and mixing duration, rate of nutrient inflow) and primary cellular metabolism (e.g., product synthesis and nutrient uptake). In this subfield of bioprocess engineering, machine learning has contributed to necessary research such as (1) estimating and predicting state variables at some points in the future (e.g., biomass concentration), (2) monitoring the factors that affect the bioreactor's performance, and (3) automating the bioprocess regarding safe operation and control purposes.
The next subfield of bioprocess engineering is Biodevices and biosensors which machine learning implementation can be found in three primary areas: (1) optimization and control of microbial fuel cells, (2) development of soft and microfluidic sensors, and (3) chemical analysis of data collected from real-time measurements.
Next, we also highlight the implementation of machine learning models to assist in the design and engineering of biomaterials in which biological engineers are interested in three primary research goals: (1) the efficient design and production of existing biological materials, (2) acceleration in developing new biological materials or improving the existing functions; and (3) quantification and automation of structural-functional relationships.
The next subfield that the authors want to summarize in this review is metabolic engineering, in which the application of machine learning focuses on (1) completing the missing information to reconstruct the metabolic network, (2) identifying essential and influential enzymes and genes expression to product synthesis, and (3) exploiting the complex interactions between omics from fluxomics to genomics and growth kinetics of extracellular microorganisms.

As mentioned in sections earlier, we highlight the bioprocess tasks, experimental datasets, and machine-learning approaches within the subfields.
Note that there are two fundamental goals when experts manipulate a bioprocess.
The first goal is to make an accurate translation from bioprocess problems to appropriate machine learning tasks that can produce a correct prediction on the experimental datasets.
The second goal is to ensure that anyone in the same laboratory or further researchers can reproduce the experiments.
Therefore, we will also provide an in-depth investigation of the reproducibility capability of these mentioned research so that we either believe in the results or build up confidence in reproducing the whole experiments and improving further from there. 

%For the ease of presentation, the abbreviation of models mentioned in Table \ref{tab:app-ml-bio-eng} is listed as follows.
Many machine learning models have been utilized and integrated into bioprocess systems are 
support vector regression (SVR), partial least square regression (PLSR), multi-gene genetic programming (MGGP), artificial neural network (ANN), Gaussian process (GP), Convolutional neural network (CNN), nonlinear model predictive control (NMPC), hierarchical recurrent sensing network (HRSN), recurrent neural network (RNN), multilayer perceptron (MLP), relevant vector machine (RVM), accelerating genetic algorithm (AGA), K-nearest neighbors (KNN), support vector machine (SVM), convolutional neural network (CNN), and principal components analysis (PCA). 
The authors do not aim to introduce and explain all the above models again, which could be referred to many references \cite{presnell2019systems,banner2021decade}.

% \usepackage{multirow}
% \usepackage{lscape}
% \usepackage{longtable}
% Note: It may be necessary to compile the document several times to get a multi-page table to line up properly
\begin{landscape}
	\begin{longtable}[c]{|l|l|l|l|ccc|c|l|}
		\hline
		\multicolumn{1}{|c|}{\multirow{2}{*}{Subfield}}                                         & \multicolumn{1}{c|}{\multirow{2}{*}{Task}}                                                                                                            & \multicolumn{1}{c|}{\multirow{2}{*}{Dataset}}                                                                                                                                                                                                                                         & \multicolumn{1}{c|}{\multirow{2}{*}{\begin{tabular}[c]{@{}c@{}}Machine learning\\ model\end{tabular}}}                                                                                                                                     & \multicolumn{3}{c|}{Reproducibility}                                             & \multirow{2}{*}{\begin{tabular}[c]{@{}c@{}}Meta\\ data\end{tabular}} & \multicolumn{1}{c|}{\multirow{2}{*}{Ref.}}         \\ \cline{5-7}
		\multicolumn{1}{|c|}{}                                                                  & \multicolumn{1}{c|}{}                                                                                                                                 & \multicolumn{1}{c|}{}                                                                                                                                                                                                                                                                 & \multicolumn{1}{c|}{}                                                                                                                                                                                                                      & \multicolumn{1}{l|}{low} & \multicolumn{1}{l|}{med.} & \multicolumn{1}{l|}{high} &                                                                      & \multicolumn{1}{c|}{}                              \\ \hline
		\endfirsthead
		\multicolumn{9}{c}%
		{{\bfseries Table \thetable\ continued from previous page}} \\
		\hline
		\multicolumn{1}{|c|}{\multirow{2}{*}{Subfield}}                                         & \multicolumn{1}{c|}{\multirow{2}{*}{Task}}                                                                                                            & \multicolumn{1}{c|}{\multirow{2}{*}{Dataset}}                                                                                                                                                                                                                                         & \multicolumn{1}{c|}{\multirow{2}{*}{\begin{tabular}[c]{@{}c@{}}Machine learning\\ model\end{tabular}}}                                                                                                                                     & \multicolumn{3}{c|}{Reproducibility}                                             & \multirow{2}{*}{\begin{tabular}[c]{@{}c@{}}Meta\\ data\end{tabular}} & \multicolumn{1}{c|}{\multirow{2}{*}{Ref.}}         \\ \cline{5-7}
		\multicolumn{1}{|c|}{}                                                                  & \multicolumn{1}{c|}{}                                                                                                                                 & \multicolumn{1}{c|}{}                                                                                                                                                                                                                                                                 & \multicolumn{1}{c|}{}                                                                                                                                                                                                                      & \multicolumn{1}{l|}{low} & \multicolumn{1}{l|}{med.} & \multicolumn{1}{l|}{high} &                                                                      & \multicolumn{1}{c|}{}                              \\ \hline
		\endhead
		\multirow{7}{*}{\begin{tabular}[c]{@{}l@{}}Bioreactor\\ engineering\end{tabular}}       & \begin{tabular}[c]{@{}l@{}}Predict the final antibody\\ and lactate concentration.\end{tabular}                                                       & \begin{tabular}[c]{@{}l@{}}Time series data of 134 temporal \\ process parameters in four seed \\ cultures (80L, 400L, 2000L, \\ 12000L). Train-test ratio 90-10. \\ 10-fold cross-validation. \\ Data are not available.\end{tabular}                                                & \begin{tabular}[c]{@{}l@{}}SVR in LIBSVM.  \\ PLSR in SIMPLS\end{tabular}                                                                                                                                                                  & \multicolumn{1}{c|}{X}   & \multicolumn{1}{c|}{}     &                           & Yes                                                                  & \cite{le2012multivariate}         \\ \cline{2-9} 
		& \begin{tabular}[c]{@{}l@{}}Predict the performance of\\ microbial  fuel cell (MFC).\end{tabular}                                                      & \begin{tabular}[c]{@{}l@{}}Data were taken from \\ \cite{wei2012study}\\ Train-test ratio 80-20. \\ Data are not available.\end{tabular}                                                                                                                             & \begin{tabular}[c]{@{}l@{}}MGGP in MATLAB R2010b \\ using GPTIPS software. \\ SVR in MATLAB R2010b \\ using LS-SVM toolbox. ANN \\ in statistical software JMP \\ version 9 (1 hidden layer, \\ 2-9 neurons in hidden layer).\end{tabular} & \multicolumn{1}{c|}{X}   & \multicolumn{1}{c|}{}     &                           & Yes                                                                  & \cite{garg2014performance}        \\ \cline{2-9} 
		& \begin{tabular}[c]{@{}l@{}}Simulate lutein \\ bioproduction process \\ control and  optimization.\end{tabular}                                        & \begin{tabular}[c]{@{}l@{}}4 sets of data, each containing 12 \\ datapoints. 50 replications of \\ artificial  datasets were produced. \\ Train-test ratio 3/4-1/4. Data are\\ not available.\end{tabular}                                                                            & \begin{tabular}[c]{@{}l@{}}ANN in pybrain library \\ (2 hidden layers, 20 neurons\\ per hidden layer).\end{tabular}                                                                                                                        & \multicolumn{1}{c|}{X}   & \multicolumn{1}{c|}{}     &                           & Yes                                                                  & \cite{del2017efficient}           \\ \cline{2-9} 
		& \begin{tabular}[c]{@{}l@{}}Predict the evolution of \\ multivariate  states for \\ lutein production process.\end{tabular}                            & \begin{tabular}[c]{@{}l@{}}No clear information about \\ experimented data. Data are \\ not available.\end{tabular}                                                                                                                                                                   & \begin{tabular}[c]{@{}l@{}}GP. Compare with 1 and 2 \\ hidden layers ANN. \\ Neurons per layer in\\ \{3,5,10,15,20,25\}.\end{tabular}                                                                                                      & \multicolumn{1}{c|}{X}   & \multicolumn{1}{c|}{}     &                           & No                                                                   & \cite{bradford2018dynamic}        \\ \cline{2-9} 
		& \begin{tabular}[c]{@{}l@{}}Simulate the fed-batch \\ production  process for \\ cyanobacterial \\ C-phycocyanin.\end{tabular}                         & \begin{tabular}[c]{@{}l@{}}Biomass concentration, nitrate\\ concentration, and phycocyanin \\ productionwere measured every \\ 8 hours. The original dataset \\ consists of 135 data  points, plus \\ 100 artificially generated data \\ points. Data are not available.\end{tabular} & \begin{tabular}[c]{@{}l@{}}ANN, no configuration \\ given.\end{tabular}                                                                                                                                                                    & \multicolumn{1}{c|}{X}   & \multicolumn{1}{c|}{}     &                           & No                                                                   & \cite{del2016dynamic}             \\ \cline{2-9} 
		& \begin{tabular}[c]{@{}l@{}}Simulate the algal biomass \\ growth and bisabolene \\ production.\end{tabular}                                            & \begin{tabular}[c]{@{}l@{}}Data were taken from 40 different \\ experimental scenarios on different\\ 120L  photobioreactors. Each \\ scenario contains 9000 data points. \\ Train-test ratio 70-30. \\ Data are not available.\end{tabular}                                          & \begin{tabular}[c]{@{}l@{}}CNN, 1 input layer with 7 \\ neurons, 2 hidden layers\\ containing convolutional\\ blocks, 1 output layer \\ with 3 neurons.\end{tabular}                                                                       & \multicolumn{1}{c|}{X}   & \multicolumn{1}{c|}{}     &                           & Yes                                                                  & \cite{del2019deep}                \\ \cline{2-9} 
		& \begin{tabular}[c]{@{}l@{}}Compare 6 different \\ GP-based NMPC models \\ for finite horizon  control.\end{tabular}                                   & \begin{tabular}[c]{@{}l@{}}Simulated dataset. Data are not\\ available.\end{tabular}                                                                                                                                                                                                  & GP, no configuration given.                                                                                                                                                                                                                & \multicolumn{1}{c|}{X}   & \multicolumn{1}{c|}{}     &                           & No                                                                   & \cite{bradford2020stochastic}     \\ \hline
		\multirow{6}{*}{\begin{tabular}[c]{@{}l@{}}Biodevices\\ and \\ biosensors\end{tabular}} & \begin{tabular}[c]{@{}l@{}}Characterize microfluidic \\ soft sensor.\end{tabular}                                                                     & \begin{tabular}[c]{@{}l@{}}Data were collected from two soft \\ pressure  sensors. The processed \\ data are available on Github.\end{tabular}                                                                                                                                        & \begin{tabular}[c]{@{}l@{}}HRSN based on RNN. \\ The model is provided\\ on Github.\end{tabular}                                                                                                                                           & \multicolumn{1}{c|}{}    & \multicolumn{1}{c|}{X}    &                           & Yes                                                                  & \cite{han2018use}                 \\ \cline{2-9} 
		& \begin{tabular}[c]{@{}l@{}}Process higher-throughput \\ Raman spectroscopy and \\ molecular images.\end{tabular}                                      & \begin{tabular}[c]{@{}l@{}}1.5 million hyperspectral Raman\\ images. The processed data are \\ available on Github.\end{tabular}                                                                                                                                                      & \begin{tabular}[c]{@{}l@{}}ResUNet. Applied transfer\\ learning technique. The \\ model is provided on \\ Github.\end{tabular}                                                                                                             & \multicolumn{1}{c|}{}    & \multicolumn{1}{c|}{X}    &                           & Yes                                                                  & \cite{horgan2021high}             \\ \cline{2-9} 
		& \begin{tabular}[c]{@{}l@{}}Diagnose anatomical \\ ex-vivo  eye tissue \\ segments in the usage of \\ Raman spectroscopy.\end{tabular}                 & \begin{tabular}[c]{@{}l@{}}Data were collected from 11 \\ separate enucleated eyes, \\ consisting of 88 spectra scans \\ per tissue segment. The \\ processed data are available\\ on Github.\end{tabular}                                                                            & \begin{tabular}[c]{@{}l@{}}Self optimising Kohonen \\ index network (SKiNET).\\ The model is provided on\\ Github.\end{tabular}                                                                                                            & \multicolumn{1}{c|}{}    & \multicolumn{1}{c|}{X}    &                           & Yes                                                                  & \cite{banbury2019development}     \\ \cline{2-9} 
		& \begin{tabular}[c]{@{}l@{}}Classify traumatic brain \\ injury severity via Raman\\ spectroscopy of the retina.\end{tabular}                           & \begin{tabular}[c]{@{}l@{}}14400 retina tissue samples were \\ collected  from adult male mice. \\ Train-test ratio 80-20. \\ 10-fold cross-validation.\\ Data are not available.\end{tabular}                                                                                        & \begin{tabular}[c]{@{}l@{}}Self optimising Kohonen\\ index network (SKiNET). \\ The model is provided on\\ Github.\end{tabular}                                                                                                            & \multicolumn{1}{c|}{X}   & \multicolumn{1}{c|}{}     &                           & Yes                                                                  & \cite{banbury2020spectroscopic}   \\ \cline{2-9} 
		& \begin{tabular}[c]{@{}l@{}}Predict bioelectricity\\ production in microbial\\ fuel cells.\end{tabular}                                                & \begin{tabular}[c]{@{}l@{}}Train-test ratio 70-30. Data are not \\ available.\end{tabular}                                                                                                                                                                                            & \begin{tabular}[c]{@{}l@{}}MLP with \{2,3,4,5\} \\ neurons. The model is \\ not available.\end{tabular}                                                                                                                                    & \multicolumn{1}{c|}{X}   & \multicolumn{1}{c|}{}     &                           & Yes                                                                  & \cite{tardast2014use}             \\ \cline{2-9} 
		& \begin{tabular}[c]{@{}l@{}}Optimize the operation of\\ multi variable microbial\\ fuel cells.\end{tabular}                                            & Data are not available.                                                                                                                                                                                                                                                               & \begin{tabular}[c]{@{}l@{}}Combination of uniform \\ design,  RVM, and AGA. \\ No configuration given.\end{tabular}                                                                                                                        & \multicolumn{1}{c|}{X}   & \multicolumn{1}{c|}{}     &                           & No                                                                   & \cite{fang2013optimizing}         \\ \hline
		\multirow{7}{*}{\begin{tabular}[c]{@{}l@{}}Biomaterials\\ engineering\end{tabular}}     & \begin{tabular}[c]{@{}l@{}}Predict phase of \\ high-entropy alloys.\end{tabular}                                                                      & \begin{tabular}[c]{@{}l@{}}Data were taken from \\ \cite{miracle2017critical}.\\ Data are not available.\end{tabular}                                                                                                                                                & \begin{tabular}[c]{@{}l@{}}KNN with k = \{1,2,...,10\}, \\ SVM in MATLAB. ANN \\ (1 input layer with 5 neurons, \\ 3 hidden layers with  5 \\ neurons each, 1 output layer\\ with 3 neurons).\end{tabular}                                 & \multicolumn{1}{c|}{X}   & \multicolumn{1}{c|}{}     &                           & No                                                                   & \cite{huang2019machine}           \\ \cline{2-9} 
		& \begin{tabular}[c]{@{}l@{}}Predict mechanical\\ functionality of protein \\ networks from confocal \\ microscopy imaging.\end{tabular}                & \begin{tabular}[c]{@{}l@{}}Data were generated in MATLAB\\ R2019  consisting of 26 calculated\\ structural  features of 37 protein\\ networks. 5-fold cross-validation. \\ Data are not available.\end{tabular}                                                                       & \begin{tabular}[c]{@{}l@{}}Gradient boosting, random \\ forest. The models are \\ provided on Github.\end{tabular}                                                                                                                         & \multicolumn{1}{c|}{X}   & \multicolumn{1}{c|}{}     &                           & No                                                                   & \cite{asgharzadeh2020nanofe}      \\ \cline{2-9} 
		& \begin{tabular}[c]{@{}l@{}}Predict the performance\\ of metal-organic \\ frameworks (MOFs).\end{tabular}                                              & \begin{tabular}[c]{@{}l@{}}3385 MOFs containing 41 distinct\\ network topologies. Processed data\\ are available.\end{tabular}                                                                                                                                                        & \begin{tabular}[c]{@{}l@{}}ANN (1 hidden layer with 30 \\ neurons). Experimental \\ reproducibility has given on a \\ dedicated website \cite{mof}.\end{tabular}                                                          & \multicolumn{1}{c|}{}    & \multicolumn{1}{c|}{X}    &                           & Yes                                                                  & \cite{moghadam2019structure}      \\ \cline{2-9} 
		& \begin{tabular}[c]{@{}l@{}}Predict injection of\\ microparticles through\\ hypodermic needles\end{tabular}                                            & \begin{tabular}[c]{@{}l@{}}Train-validation-test ratio 60-20-20.\\ Raw data and codes can be provided\\ upon reasonable request. Licensing\\ fees might be applied.\end{tabular}                                                                                                      & \begin{tabular}[c]{@{}l@{}}ANN (10 hidden layers)\\ in MATLAB Deep\\ Learning Toolbox. No\\ configuration given.\end{tabular}                                                                                                              & \multicolumn{1}{c|}{X}   & \multicolumn{1}{c|}{}     &                           & Yes                                                                  & \cite{sarmadi2020modeling}        \\ \cline{2-9} 
		& \begin{tabular}[c]{@{}l@{}}Detect amino acids with\\ nanoporous single-layer\\ molybdenum disulfide.\end{tabular}                                     & \begin{tabular}[c]{@{}l@{}}Raw data and codes can be \\ provided upon reasonable\\ request.\end{tabular}                                                                                                                                                                              & \begin{tabular}[c]{@{}l@{}}KNN, random forest, logistic\\ regression. No configuration\\ given.\end{tabular}                                                                                                                               & \multicolumn{1}{c|}{X}   & \multicolumn{1}{c|}{}     &                           & No                                                                   & \cite{farimani2018identification} \\ \cline{2-9} 
		& \begin{tabular}[c]{@{}l@{}}Classify cell shape\\ phenotypes.\end{tabular}                                                                             & Data are not available.                                                                                                                                                                                                                                                               & \begin{tabular}[c]{@{}l@{}}SVM in MATLAB. No\\ configuration given.\end{tabular}                                                                                                                                                           & \multicolumn{1}{c|}{X}   & \multicolumn{1}{c|}{}     &                           & Yes                                                                  & \cite{tourlomousis2019machine}    \\ \cline{2-9} 
		& \begin{tabular}[c]{@{}l@{}}Detect scattering effect\\ in light-based 3D printing.\end{tabular}                                                        & \begin{tabular}[c]{@{}l@{}}300 mask-structure pairs plus 600\\ pairs augmented. Data are not\\ available.\end{tabular}                                                                                                                                                                & \begin{tabular}[c]{@{}l@{}}Neural network with 14-layer\\ CNN architecture combined\\ with U-Net skip connections.\\ The model is not available.\end{tabular}                                                                              & \multicolumn{1}{c|}{X}   & \multicolumn{1}{c|}{}     &                           & Yes                                                                  & \cite{you2020mitigating}          \\ \hline
		\multirow{9}{*}{\begin{tabular}[c]{@{}l@{}}Metabolic\\ engineering\end{tabular}}        & \begin{tabular}[c]{@{}l@{}}Fill gaps in a metabolic\\ network.\end{tabular}                                                                           & \begin{tabular}[c]{@{}l@{}}BiGG database \cite{king2016bigg}.\\ Data are available on \\ BoostGapFill's Github.\end{tabular}                                                                                                                                         & \begin{tabular}[c]{@{}l@{}}BoostGapFill open source\\ tool.\end{tabular}                                                                                                                                                                   & \multicolumn{1}{c|}{}    & \multicolumn{1}{c|}{}     & X                         & Yes                                                                  & \cite{oyetunde2017boostgapfill}   \\ \cline{2-9} 
		& \begin{tabular}[c]{@{}l@{}}Identify specific enzymes\\ limiting production in a\\ pathway.\end{tabular}                                               & Data are not available.                                                                                                                                                                                                                                                               & PCA in MATLAB.                                                                                                                                                                                                                             & \multicolumn{1}{c|}{X}   & \multicolumn{1}{c|}{}     &                           & Yes                                                                  & \cite{alonso2015principal}        \\ \cline{2-9} 
		& \begin{tabular}[c]{@{}l@{}}Predict the bacterial\\ central metabolism.\end{tabular}                                                                   & \begin{tabular}[c]{@{}l@{}}Data collected from 100 \\ C-metabolic flux analysis \\ papers. Data are available.\end{tabular}                                                                                                                                                           & \begin{tabular}[c]{@{}l@{}}MFlux web-based platform.\\ Source codes are available.\\ MFlux applies SVM, KNN, \\ decision tree.\end{tabular}                                                                                                & \multicolumn{1}{c|}{}    & \multicolumn{1}{c|}{}     & X                         & Yes                                                                  & \cite{wu2016rapid}                \\ \cline{2-9} 
		& \begin{tabular}[c]{@{}l@{}}Predict essential genes in\\ Escherichia coli \\ metabolism.\end{tabular}                                                  & \begin{tabular}[c]{@{}l@{}}4094 metabolic reaction-gene pairs. \\ Several  additional datasets from \\ private providers and E. coli Gene \\ Expression Database. Data are\\ not available.\end{tabular}                                                                              & \begin{tabular}[c]{@{}l@{}}SVM. No configuration\\ given.\end{tabular}                                                                                                                                                                     & \multicolumn{1}{c|}{X}   & \multicolumn{1}{c|}{}     &                           & Yes                                                                  & \cite{nandi2017integrative}       \\ \cline{2-9} 
		& \begin{tabular}[c]{@{}l@{}}Selecting substrates that\\ best expand an enzyme’s \\ promiscuity.\end{tabular}                                           & \begin{tabular}[c]{@{}l@{}}BRENDA online enzyme database\\ \cite{schomburg2004brenda}.\end{tabular}                                                                                                                                                                  & \begin{tabular}[c]{@{}l@{}}SVM. Apply active learning\\ approach.\end{tabular}                                                                                                                                                             & \multicolumn{1}{c|}{}    & \multicolumn{1}{c|}{X}    &                           & Yes                                                                  & \cite{liu2015genome}              \\ \cline{2-9} 
		& \begin{tabular}[c]{@{}l@{}}Propose a real-time \\ optimization for the \\ control of co-cultures \\ within the continuous\\ bioreactors.\end{tabular} & \begin{tabular}[c]{@{}l@{}}Simulated data of continuous\\ bioreactor, 24h duration, \\ measurement every 5 min. \\ Data is not available.\end{tabular}                                                                                                                                & \begin{tabular}[c]{@{}l@{}}Neural fitted Q-learning, \\ reinforcement learning. The\\ models are provided on\\ Github.\end{tabular}                                                                                                        & \multicolumn{1}{l|}{X}   & \multicolumn{1}{l|}{}     & \multicolumn{1}{l|}{}     & \multicolumn{1}{l|}{No}                                              & \cite{treloar2020deep}            \\ \cline{2-9} 
		& \begin{tabular}[c]{@{}l@{}}Predict xylose \\ consumption, biomass\\ and xylitol production.\end{tabular}                                              & \begin{tabular}[c]{@{}l@{}}Datasets were collected every 6 h \\ interval resulting in 340 data points \\ (27 runs). Data are not available.\end{tabular}                                                                                                                              & \begin{tabular}[c]{@{}l@{}}ANN of 5-10-2 topology in\\ MATLAB Deep Learning\\ Toolbox.\end{tabular}                                                                                                                                        & \multicolumn{1}{l|}{X}   & \multicolumn{1}{l|}{}     & \multicolumn{1}{l|}{}     & \multicolumn{1}{l|}{Yes}                                             & \cite{pappu2017artificial}        \\ \cline{2-9} 
		& \begin{tabular}[c]{@{}l@{}}Explore the \\ bioretrosynthesis\\ space in synthetic \\ pathway design.\end{tabular}                                      & \begin{tabular}[c]{@{}l@{}}Golden dataset of 20 manually\\ curated experimental pathways, \\ 152 metabolic engineering \\ projects. Data are available.\end{tabular}                                                                                                                  & \begin{tabular}[c]{@{}l@{}}Applied reinforcement \\ learning. The open source\\ solution is provided on \\ Github.\end{tabular}                                                                                                            & \multicolumn{1}{l|}{}    & \multicolumn{1}{l|}{}     & \multicolumn{1}{l|}{X}    & \multicolumn{1}{l|}{Yes}                                             & \cite{koch2019reinforcement}      \\ \cline{2-9} 
		& \begin{tabular}[c]{@{}l@{}}Predict protein expression\\ from promoter sequences.\end{tabular}                                                         & \begin{tabular}[c]{@{}l@{}}675,000 constitutive and 327,000\\ inducible promoter sequences.\\ Data are available.\end{tabular}                                                                                                                                                        & \begin{tabular}[c]{@{}l@{}}CNN, no configuration\\ given. The model is\\ available on Github.\end{tabular}                                                                                                                                 & \multicolumn{1}{l|}{}    & \multicolumn{1}{l|}{}     & \multicolumn{1}{l|}{X}    & \multicolumn{1}{l|}{Yes}                                             & \cite{kotopka2020model}           \\ \hline
		\caption{Application of machine learning in various subfields of bioprocess engineering.}
		\label{tab:app-ml-bio-eng}\\
	\end{longtable}
\end{landscape}

Table \ref{tab:app-ml-bio-eng} shows us a huge problem with the information needed to ensure the reliability and reproducibility of the experiment. 
When we sorted by problem requirements, data, and machine learning models from a machine learning perspective, we witness more clearly the inadequacies and inconsistencies in the information presented in those published studies. 
Metadata about curation, provenance, and aggregation is not clearly described. 
The infrastructure condition and data engineering pipeline are not mentioned. 
Those papers do not record the model construction process or justification for design decisions. 
It leads to the acceptance of simplifications and assumptions about the system design, environmental context, biological or biochemical features, and other artifacts. 
Note that the required information will help avoid unnecessary recreation and model version control. 
A number of data and models in bioengineering are conducted from a simulation process. 
It is the best practice if all simulation inputs and applied methods, initial conditions, numerical integration algorithms, seed values, and other emerged data should be carefully recorded. 
If any parameters and hyperparameters are estimated, make sure to share the estimation algorithms and the value ranges. 
As the results of our mentioned issues, Table \ref{tab:app-ml-bio-eng} is not consistent in how it describes the case studies as in some cases the programming language used is given but not in all. 
All columns in this figure should contain the same amount of information to allow the readers to compare between studies. 
How can we collect the required amount of information if they are not provided? 
It is a clear confirmation of the reproducibility problem that we want to discuss in the paper.

%% file: tex/Challenges.tex
%\section{Challenges and Future Research Directions}

\subsection{Challenge 1: Reproducibility Crisis}

Machine learning is, to a considerable extent, an experimental science. 
As a result, the reproducibility of computational pipelines is of significant concern \cite{hutson2018artificial,gundersen2018state,haibe2020transparency}. 
Machine learning experts have highlighted that the reproducibility of scientific results is a key
element of science and credibility of conclusions made to the extent that they explicitly encourage replicating the experimental results of any published study \cite {bouthillier2019unreproducible, pineau2021improving, bouthillier2020survey, leipzig2021role, alahmari2020challenges}. 
In the nine major machine learning conferences, including NeurIPS, ICML, ICLR, ACL-IJCNLP, EMNLP, CVPR, ICCV, AAAI, and IJCAI, the criterion of reproducibility has been highly required in every peer-reviewed process and published research paper \cite{raff2019step,sethi2018dlpaper2code}. 
To establish which algorithm is better for a learning task, it is an essential rule that any computational experiment for algorithm assessment should be carried out on the same datasets representing the task. 
This dataset must be publicly available or published together with the first paper addressing this task. 
The evaluation metrics will be calculated using the same formulas as the first published paper. 
In the case of using a new set of formulas, it is necessary to re-test the model in the first publication, applying the methods of optimal search for the participants on this new set of formulas. 
Take an example as follows, we have two algorithms to compare. 
Algorithm A is our development, and algorithm B is proposed by previous research. 
The comparison results depend on how much documentation is publicly made available.
For example, if we only have access to the written documents as published articles, we have to self-implement algorithm B and test it on the data we collect ourselves. 
In fact, there is practically no way to verify that we have implemented and configured the algorithm in precisely the same way as the original authors, especially the values used for hyperparameters. 
Thus, the more literature (articles, algorithmic code, and data) provided by the original authors, the easier it is for independent researchers to reproduce and demonstrate the published results that the claims made are valid. 
We proceed with the problem further regarding the above algorithms A and B. 
Suppose we want to test algorithm A on the same published data set. In that case, the question is whether we have to test algorithm B again to verify the correctness or accept the results reported as comparative results. 
This is a relevant question because newly proposed algorithms are often compared with published models developed by third parties without re-testing. 
However, one scenario exists when algorithm B compares itself with many previous algorithms, but the code is not publicly available. 
And instead, later researchers often take reported results to compare and accept as proven facts. 
In addition, independent research experts have found it difficult to obtain similar results when re-implementing complete experiments reported in the scientific literature if the values for some important parameters and hyperparameters are not given. 
Computer science, specifically machine learning, is in a favorable situation where identical empirical procedures can be followed using the same data sets.
Although in this case, the biggest challenge is the lab, different hardware, and software where the experiments are conducted. 
Reproducibility is best demonstrated by applying algorithms A and B on the same data but for different laboratory, configurations to produce similar results and arrive at the same conclusions.
Interest has grown not only in the machine learning community but also in bioengineering \cite{jessop2019improving,amanullah2010novel,fuchs2021newly}, biomedical engineering \cite{raphael2020controlled}, biology \cite{roper2022testing}, and genome editing \cite{teboul2020variability} regarding the reproducibility of published scientific results. 

However, the reproducibility requirement for biological systems is much more difficult because data are extracted from living organisms, chemicals, and organic interactions, e.g., proteins and strains of cells. 
In systems biology modeling, the issue of reproducibility involves a combination of not having FAIR experimental data and difficult-to-reproduce model fitting strategies due to missing parameters, initial conditions, and inconsistent model structure \cite{tiwari2021reproducibility}.
Even the biomass collected during experiments in the same laboratory, on the same bioreactor, differed by the time of year or collected by different technicians. 
This complicates efforts to apply approaches from the field of machine learning, where data is more stable and redundant. 
Furthermore, increasingly sophisticated bioengineering tools are making cell biology experiments more complex. 
The time to conduct biological experiments is also longer, leading to more complex reproducibility. 
Thorough validation can take months or even years to complete. That makes it difficult for laboratories that are not equipped with modern equipment to reproduce experimental conditions that more qualified laboratories have done. 
Instead, the biological sciences depend on other less reliable techniques for reproducing experiments, resulting in publications that are less conditional on comparison with previous studies which makes data difficult to share or reuse.

According to a Nature survey of 1576 researchers, M. Baker points out that the scientific community has a general view that there is an ongoing reproducibility crisis \cite{baker20161}. 
Surveys have shown that more than 70\% of researchers have tried and failed to reproduce other scientists' experiments, more than 50\% have been unable to replicate their own experiments, and more than 30\% believe in published results even though they acknowledge that published results may be wrong. 
Another interesting survey published on Molecular Systems Biology\footnote{\url{https://www.ebi.ac.uk/biomodels/reproducibility}} \cite{tiwari2021reproducibility} that the authors investigated 455 kinetic models of various biological processes.
The authors concluded that 49\% of the models could not be reproduced using the information provided in the manuscripts.
They even proceeded with an effort by contacting and asking the authors of 455 published papers.
And surprisingly, only 12\% out of 49\% could be reproduced.
The plausible reasons for non-reproducibility include inconsistency in model structure, missing parameter values, missing initial concentration, and even unknown reasons.
Many bioengineering professionals reuse machine learning as a complete implementation on a particular computing platform. 
However, another study even concluded that a machine learning platform does not guarantee computational reproducibility and that the test results generated from a machine learning platform cannot be trusted entirely \cite {gundersen2022machine}.

\subsection{Proposed Research 1: Promote a Culture of Inferential Reproducibility in Bioengineering}
\begin{figure}[ht!]
	\centering
	\includegraphics[width=1\textwidth]{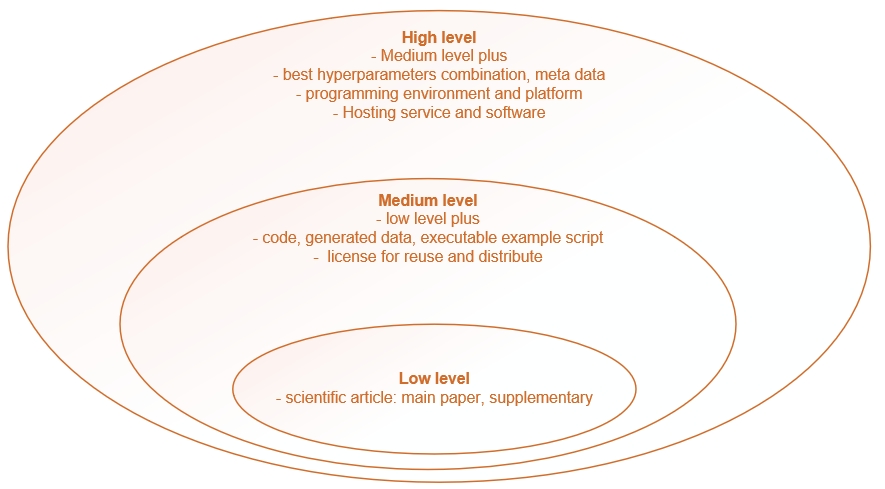}
	\caption{Reproducibility levels from the lowest, e.g., Low Level, to the highest, e.g., High Level. Each level requires some performance and proof.}
	\label{fig:reproducibility}
\end{figure}

In addition to the techniques and methodologies proposed in machine learning \cite{goodman2016does,dirnagl2019rethinking,tatman2018practical}, we need to change the culture regarding research reproducibility. 
We must encourage the practice of reproducibility and help subsequent researchers to enforce it as a cornerstone of science \cite{porubsky2020best}.
Reproducible models confer essential benefits because they are easier to understand, trust, modify and reuse.
This facilitates our collaboration better and is more open, thus, attracting follow-up studies to construct multi-scale models of larger, more complex systems from the current results. 
We need to fund and encourage individuals and research groups to confirm (or sometimes disprove) the findings of others with reproducible results. 
We should not criticize studies whose findings cannot be confirmed. 
In contrast, our work attempts to replicate highly reproducible studies, even if the results are not precisely the same. 
Journals can even create a new criteria category for assessing which research supports or integrates research reproducibility. 
The study replication levels can be found in Figure \ref{fig:reproducibility}.
The lowest level of reproducibility requires a research article and possible supplementary where the researchers should describe how bioprocess experiments have been conducted.
Other metadata should also be available. 
However, the experimental codes and datasets, or the executable scripts can be missing.
Hence, by fulfilling those mentioned missing codes, datasets, and scripts, the reproducibility is improved to the Medium level.
The High level requires basic machine learning in production where the programming environment, a platform of development, hosting and metadata are presented.
%Hence, we proposed a checklist to access each model and address the reproducibility levels.

\subsection{Challenge 2: Benchmark Datasets and Evaluations of Bioengineering approaches}
Within computer science, benchmarking is the development of guidelines and best practices. 
It contains three sub-fields: scientific machine learning benchmark, application benchmark, and system benchmark \cite{thiyagalingam2022scientific}. 
Application benchmark concerns the complete deployment of machine learning applications using various hardware and software settings. 
The benchmark evaluates the use of resources, e.g., file systems, software libraries and versions, hardware configuration, and scaling factor of computing capacity, that affect the time-to-solution of the application. 
System benchmark concentrates on the availability of a machine learning application in a broader environment. 
The system benchmark evaluates network throughput and the number of floating-point operations per second. 
These two benchmarking frameworks are not technically suitable for bioprocess engineering. 
This article focuses on the machine learning benchmark and how to promote it within bioprocess engineering research. 
The machine learning benchmark is much simpler and easy to implement as it requires two subjects: datasets and reference models
%as presented in Figure \ref{fig:ml-benchmark}. 
Firstly, benchmark datasets which are the fundamental cornerstone of machine learning should be made available to the research community. 
The exact training, validation, and test sets are on which all the reference implementation must be based on. 
Secondly, proposed approaches and state-of-the-art modeling applications will be developed over time and considered blueprints for further use on different benchmark datasets.
Thirdly, an excellent overall design of the machine learning benchmark has fostered great boosters for research and discussion of corresponding areas: out-of-the-box downloading and usage, interoperability, and ease of customization \cite{zoller2021benchmark,denton2020bringing}. 
Bioprocess engineers must be able to understand the most suitable machine learning models by looking at the benchmarking performance on the equivalent datasets and types of investigated problems.
More specifically, Bioprocess experts might refer a blossoming of benchmarks on neural networks and applications \cite{dong2017dnnmark,alzahrani2021comprehensive,dwivedi2020benchmarking,hirose2021hpo,zhu2018benchmarking,sharan2021benchmarking}, time series \cite{xie2018benchmark,javed2020benchmark,fauvel2020performance,hao2021ts,bauer2021libra}, image data \cite{menze2014multimodal,russakovsky2015imagenet,lin2014microsoft}, text-based source \cite{partalas2015lshtc,bojar2014findings,rajpurkar2016squad}, via community competition \footnote{\url{https://www.kaggle.com/competitions}}, \footnote{\url{https://paperswithcode.com/}}, and many others \cite{olson2017pmlb,romano2022pmlb}.

%\begin{figure}[ht!]
%	\centering
%	\includegraphics[width=0.7\textwidth]{fig/Pic-1959.jpg}
%	\caption{Components of machine learning benchmark.}
%	\label{fig:ml-benchmark}
%\end{figure}

\subsection{Proposed Research 2: Comprehensive Construction of Bioprocess Engineering Benchmark}
The development of standards for bioprocessing engineering is essential to accelerate its growth while also attracting the participation of experts from many other fields. 
As we discussed above and the lessons learned from the machine learning community for the necessity of an ideal benchmark.
More specifically, the benchmark (1) should provide publicly available datasets, while also providing standard procedures on those public datasets like typical machine learning tasks, such as classification, regression, and prediction; and (2) must be generic enough and easily integrated to accommodate different bio-research engineering pipelines. 
However, an important point that makes the bioprocessing specification more prominent and specific to its field is the bioprocessing metadata \cite{charaniya2008mining,grover2018mining,rommel2004data,alford2006bioprocess}. 
Take a look at the following example of experimental verification. 

A laboratory, named A, performed a verification experiment of four optimally designed experiments (two were performed by a kinetic model and the others by an artificial neural network) \cite{del2019comparison}. 
These experiments were performed in a glass tubular photoreactor with a capacity of 1 L (length of 15.5 cm and diameter of 9.5 cm). 
A technician attaches an artificial light source to opposite sides of the reactor using 14 W TL 5 tungsten incandescent lamps, manufactured by Philip Co., China). 
The experiments started with two hyperparameters of biomass concentration and nitrate concentration set to 0.27 g/L and 9 mM, respectively. 
The experiments also set two other hyperparameters, the influential nitrate concentration, and the fixed culture temperature of 0.1 M and 35 °C for all experiments. 
Cultures were continuously aerated with 2.5\% CO2 in air at 0.2 vvm and pH = 7.5 at a stirring rate of 300 rpm. 
The technician varied the nitrate feeding rate and light intensity daily throughout the experiment.

\begin{figure}[ht!]
	\centering
	\includegraphics[width=0.7\textwidth]{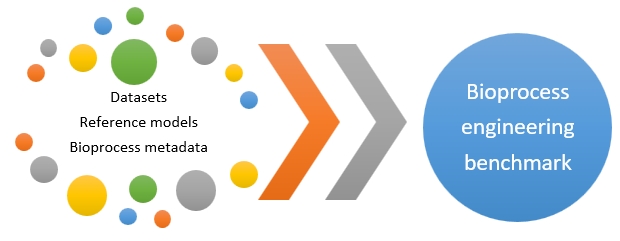}
	\caption{Components of bioprocess engineering benchmark.}
	\label{fig:ml-bio-benchmark}
\end{figure}

Let's assume that laboratory A releases the experimental datasets and reference model, e.g., an artificial neural network in this case.
If the laboratory, named B, is interested in the experiments and wants to improve its current project with a similar verification experiment. 
Then laboratory B must know the exact experimental settings and configuration such as the equipment, chemical origin, spacial location of equipment installation, nitrate feeding rate log, and other necessary metadata.
Hence, the bioprocess engineering benchmark should have the third component: bioprocess metadata as presented in Figure \ref{fig:ml-bio-benchmark}. 
Unfortunately, the bioprocess engineering literature witnesses not many the variety of benchmark-ready published articles and dedicated benchmarking \cite{wu2018moleculenet,villaverde2015biopredyn,villaverde2019benchmarking,ballnus2017comprehensive,riordon2019deep}. 

%\subsection{Challenge 3: Knowledge-based Selection of Bioprocess Hyperparameters for Machine Learning Optimization}

%\myBlue{Could bio-experts write me this section? Or should we need it?}

%\paragraph{Proposed Research 3: Automatic Pipeline of Bioprocess Hyperparameters Selection}

%% file: tex/Conclusion.tex
%\section{Conclusion}

Bioprocessing engineering is involved in solving many of the world's greatest challenges, where ML integration has proven immensely useful in bioprocess automation and reducing uncertainty in the decision-making process. 
However, many barriers and challenges need to be recognized to enhance the impact of the bio-industry. The authors confirm that the increasing use of ML in bioengineering and bioprocess automation will continue to be accelerated soon. 
This combination has been enabled by important technological advances, hardware, and software, which are continuing to evolve. 
Automation in experimental biotechnology, both at individual laboratories and the level of interlaboratory biosynthesis, coupled with robust decision-making, will be driven by better machine learning models. 
It could spark the emergence of fully automated, continuous bioprocessing techniques that do not depend on or limit human intervention. Most of the success has come from applying ML approaches developed in other domains directly to biological data sources. 
In this work, we review the existing applications of machine learning and artificial intelligence in bioprocess engineerning automation from the perspective of making a faster and less costly development spiral. 
We mostly focus on the existing tools that, in our opinion, have not been considered yet despite having a great potential to automate model-building in the different stages to make the computational methods reproducible, which in turn provides full provenance to decisions taken.
We have summarized recent ML implementation across several important subfields of bioprocess systems and raise two crucial challenges remaining the bottleneck of bioprocess automation and reducing uncertainty in biotechnology development. 
Although it is an ambitious goal, the combination of bioprocess engineering and machine learning is likely to yield many of the biggest developments in bioprocess in the coming years. 
Broader Impact: over the next decade: making bioprocess development comprehensively FAIR \cite{grandcolas2019rise,jacobsen2020fair} - Findable, Accessible, Interoperable, and Reusable.